\documentclass{article}

\PassOptionsToPackage{numbers, compress}{natbib}
\usepackage[preprint]{neurips_2026}

\usepackage[utf8]{inputenc} 
\usepackage[T1]{fontenc}    
\usepackage{hyperref}       
\usepackage{url}            
\usepackage{booktabs}       
\usepackage{amsfonts}       
\usepackage{nicefrac}       
\usepackage{microtype}      
\usepackage{xcolor}         
\usepackage{amsmath}
\usepackage{amsthm}
\usepackage{graphicx}
\usepackage{enumitem}
\usepackage{algorithm}
\usepackage{subfigure}
\usepackage{wrapfig}
\usepackage{bbm}
\usepackage{makecell}
\usepackage{algpseudocode}
\algrenewcommand\algorithmicrequire{\textbf{Input:}}
\algrenewcommand\algorithmicensure{\textbf{Output:}}
\theoremstyle{plain}
\newtheorem{theorem}{Theorem}[section]

\theoremstyle{definition}

\theoremstyle{remark}
\newtheorem{remark}[theorem]{Remark}

\definecolor{backgroundorange}{HTML}{E6C38A}
\definecolor{backgroundred}{HTML}{E6B8AE}
\definecolor{backgroundteal}{HTML}{A8C7C0}

\newcommand{\mathhl}[2][green!15]{%
  \begingroup
  \setlength{\fboxsep}{0.2pt}%
  \colorbox{#1}{$\displaystyle #2$}%
  \endgroup
}

\newcommand{\texthl}[2][backgroundred]{%
  {\setlength{\fboxsep}{0.2pt}%
   \colorbox{#1}{\strut #2}}%
}

\newcommand{\greencheck}{{\color{green}\checkmark}} 
\newcommand{\redcross}{{\color{red}$\times$}} 

\title{Boosting Reinforcement Learning with Verifiable Rewards via Randomly Selected Few-Shot Guidance}

\author{%
  Kai Yan \qquad Alexander G. Schwing \qquad Yu-Xiong Wang\\
  University of Illinois Urbana-Champaign\\
  \texttt{\{kaiyan3, aschwing, yxw\}@illinois.edu} \\
\url{https://github.com/KaiYan289/FEST}
}

\begin{document}

\maketitle

\begin{abstract}
  Reinforcement Learning with Verifiable Rewards (RLVR) has achieved great success in developing Large Language Models (LLMs) with chain-of-thought rollouts for many tasks such as math and coding. Nevertheless, RLVR struggles with sample efficiency on difficult problems where correct rollouts are hard to generate. Prior works propose to address this issue via demonstration-guided RLVR, i.e., to conduct Supervised FineTuning (SFT) when RL fails; however, SFT often requires a lot of data, which can be expensive to acquire. In this paper, we propose FEST, a \textbf{FE}w-\textbf{S}ho\textbf{T} demonstration-guided RLVR algorithm. It attains compelling results with only $128$ demonstrations randomly selected from an SFT dataset. We find that three components are vital for the success: supervised signal, on-policy signal, and decaying weights on the few-shot SFT dataset to prevent overfitting from multiple-epoch training. On several benchmarks, FEST outperforms baselines with magnitudes less SFT data, even matching their performance with full dataset. 
\end{abstract}

\section{Introduction}

Two years after Reinforcement Learning from Human Feedback (RLHF) and GPT-3.5~\cite{ouyang2022training} propelled Large Language Models (LLMs) to the forefront of AI research~\cite{ke2025survey}, a new RL paradigm has emerged. Following OpenAI o1~\cite{jaech2024openai} and DeepSeek-R1~\cite{Guo2025DeepSeek}, Reinforcement Learning with Verifiable Rewards (RLVR)~\cite{Guo2025DeepSeek} has become the second dominant RL paradigm in the community. Unlike RLHF, which assigns rewards based on subjective and often vague human preferences~\cite{wang2024secrets}, RLVR leverages objective, verifiable rewards—such as unit tests for coding~\cite{jiang2025coderlplus} or ground-truth comparisons for mathematics~\cite{cobbe2021training}. Consequently, RLVR is exceptionally well-suited for reasoning-heavy tasks. Driven by this approach, state-of-the-art LLMs have attained gold-medal performance in international competitions~\cite{huang2025winning} and are beginning to tackle open problems at the frontiers of human knowledge~\cite{open}.


Despite its impressive performance, RLVR is beset by a long-standing challenge in reinforcement learning: low sample efficiency on complex tasks. While prior knowledge from pre-training and Supervised Fine-Tuning (SFT) can partially mitigate this—analogous to imitation learning~\cite{attia2018global}—RL-trained LLMs still struggle to explore beyond the capabilities of their base models~\cite{yue2025does}. For instance, in mathematical tasks with binary rewards, a batch that fails to yield a single correct answer results in an advantage of $0$, providing no learning signal. To address this, state-of-the-art algorithms such as DAPO~\cite{yu2025dapo} and CISPO~\cite{chen2025minimax} employ repeated sampling until a success is found, which can triple the computational overhead on average~\cite{zhang2025improving}.


\begin{figure}[t]
    \centering
    \includegraphics[width=0.8\linewidth]{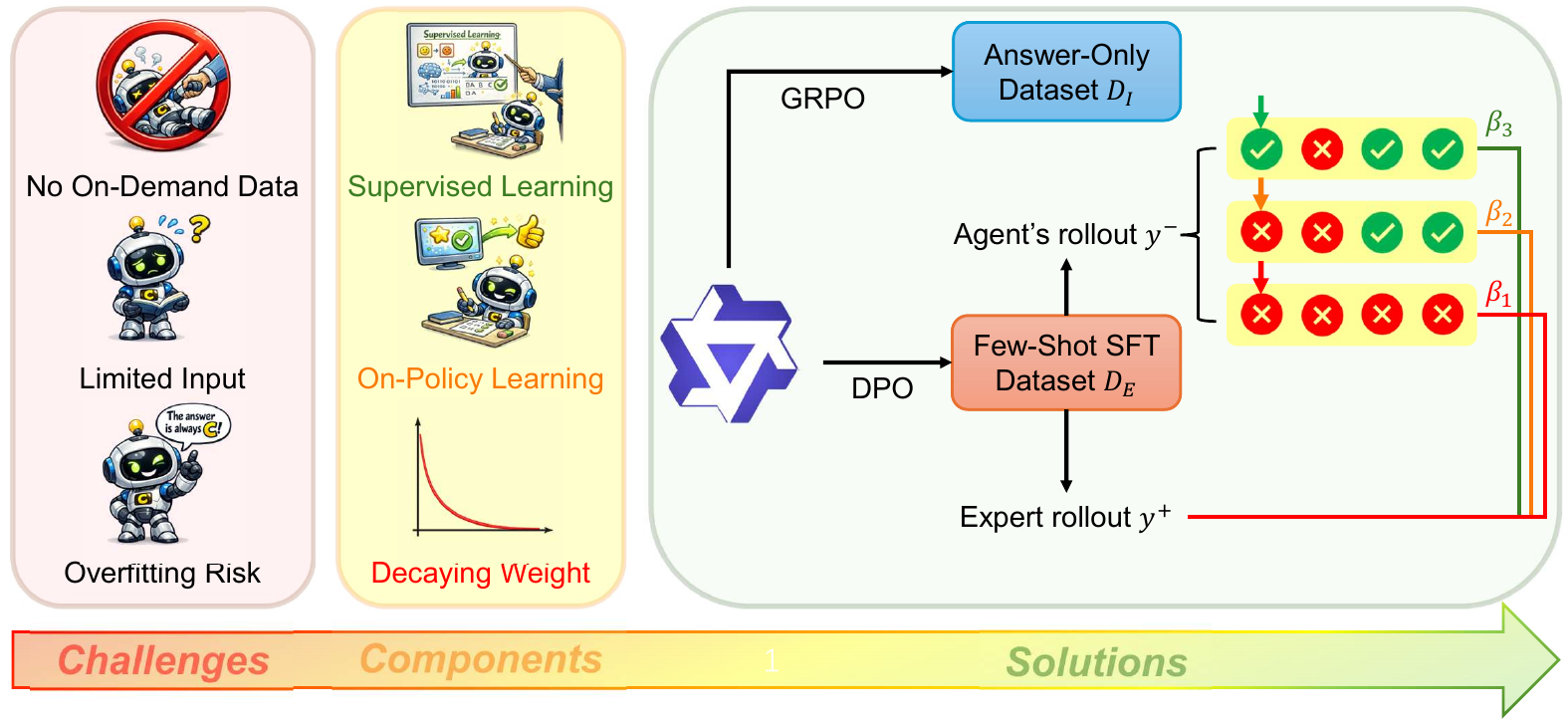}
    \caption{Overview of our work. We introduce a few-shot demonstration-guided RLVR paradigm to address three primary challenges: the lack of on-demand data, limited expert input, and high overfitting risk. To mitigate, FEST incorporates three vital components---\texthl[backgroundteal]{supervised learning}, \texthl[backgroundorange]{on-policy learning}, and \texthl[backgroundred]{decaying weight}---to effectively leverage few-shot data. Specifically, FEST jointly optimizes the model using GRPO on the answer-only data ($D_I$) and a semi-online DPO loss on the few-shot SFT data ($D_E$). The latter utilizes adaptive $\beta$ coefficients to weight different types of rollouts, bridging the gap between expert demonstrations ($y^+$) and agent-generated samples ($y^-$).}
    \label{fig:teaser}
\end{figure}

\begin{table}[t]
    \centering
    \small
    \caption{Comparison of FEST with existing demonstration-guided RLVR baselines. SFT data requirements for HPT and SuperRL are estimated through replication (with SuperRL figures derived from training on OpenR1-220K); for ReLIFT, SASR, MIFO, and CHORD, we utilize the ratios reported in their respective papers. Notably, while other methods address competition-level problems, SASR focuses on simpler tasks. We omit DyME~\cite{koksal2025few} due to its similarity to HPT and its focus on vision-based tasks rather than mathematical reasoning.}
    \begin{tabular}{cccc}
        \toprule 
        RLVR Method & SFT Data Used & SFT Data Prepared & Random SFT Data?\\
        \midrule
        SuperRL~\cite{liu2025superrl} & Approx. 50K & 220K & \redcross \\
        LUFFY~\cite{yan2025luffy} & 46K & 46K & \redcross  \\
        SRFT~\cite{fu2025srft} & 46K & 46K & \redcross \\ 
        HPT~\cite{lv2025hpt} & 10K & 46K & \redcross \\
        ReLIFT~\cite{ma2025relift} & 8.6K & 46K & \redcross \\
        MIFO~\cite{yuan2025mitigating} & 6.4K & 46K & \redcross \\
        CHORD~\cite{zhang2025onpolicy} & 5K & 5K & \greencheck \\
        SASR~\cite{chen2025sasr} & 2K$^*$ & 4K$^*$ & \redcross \\
        \textbf{FEST (Ours)} &  \textbf{128} & \textbf{128} & \greencheck \\
        \bottomrule
    \end{tabular}
    \label{tab:teaser}
\end{table}

To address these challenges, recent research proposes a novel paradigm known as demonstration-guided RL, or unified post-training~\cite{yan2025luffy}. In this framework, SFT is integrated with RL, particularly when RL sampling fails to produce positive rollouts~\cite{lv2025hpt,ma2025relift}. However, these approaches demand a large volume of SFT data, which can be prohibitively expensive~\cite{wsj_mercor_2026}. For instance, curating just 2,500 questions for ``Humanity's Last Exam''~\cite{phan2025humanity} required \textit{1,000 graduate-degree holders}, even when providing only brief rationales—a level of effort that scales poorly for training data involving long reasoning traces. While bootstrapping and distilling from existing models is an alternative~\cite{comanici2025gemini}, such practices raise significant concerns regarding legality~\cite{dornis2025generative}, proprietary API costs~\cite{irugalbandara2024scaling}, and potential model collapse~\cite{shumailov2024modelcollapse}. In contrast, answer-only RL data are more accessible; large-scale mathematical Q\&A pairs can be mined from online forums, whereas raw community answers typically necessitate extensive filtering or rewriting before serving as high-quality SFT demonstrations~\cite{zhang2024stackmathqa,mahdavi2025leveraging,albalak2025big}.


In this paper, we propose FEST, a \textit{FEw-ShoT demonstration-guided RLVR algorithm} designed to thrive with as few as 128 randomly selected examples from an SFT dataset. To extract substantial performance gains from such a limited, uncurated dataset, we identify three critical components: (i) a \texthl[backgroundteal]{supervised learning signal} to provide expert guidance; (ii) an \texthl[backgroundorange]{on-policy signal} to mitigate exposure bias and serve as adversarial training~\cite{Bengio2015ScheduledSampling} for enhanced robustness; and (iii) a \texthl[backgroundred]{decaying weight} to prevent overfitting. We find that incorporating a semi-online Direct Preference Optimization (DPO)~\cite{rafailov2023direct,chen2024self,tan2024sopo} loss—where demonstrations serve as positive examples and agent rollouts as negative ones—satisfies all three requirements. Furthermore, while our primary RL framework, Group Relative Policy Optimization (GRPO)~\cite{Guo2025DeepSeek}, operates on token-level log probabilities, standard DPO relies on sequence-level values. To resolve this mismatch, we introduce a variant of FEST that decomposes the DPO loss, replacing the online component with a GRPO loss featuring negative advantages, as supported by prior work~\cite{zhu2025surprising}. Our framework is illustrated in Fig.~\ref{fig:teaser}.


Our contributions are summarized as follows: (i) We introduce a novel post-training paradigm, few-shot demonstration-guided RLVR, which significantly boosts RLVR performance with minimal SFT data (see Tab.~\ref{tab:teaser} for a comparison with existing methods). (ii) We develop FEST by elucidating and integrating three vital components essential for this few-shot training regime. (iii) Theoretically, we extend the unified post-training framework of HPT~\cite{lv2025hpt} by incorporating DPO (Appendix~\ref{sec:hpt}). (iv) We empirically validate our approach across multiple benchmarks, demonstrating that FEST consistently outperforms various strong baselines.


\section{Preliminaries}
\label{sec:prelim}

\textbf{Reinforcement Learning with Verifiable Rewards (RLVR).} RLVR is an emerging post-training framework that facilitates solving tasks with objective ground truths, such as mathematics and programming, by eliciting complex Chain-of-Thought (CoT) reasoning~\cite{wei2022chain}. For a given prompt $x$ representing the input task\footnote{Unless otherwise noted, the RLVR settings discussed in this paper are single-turn interactions with sequence-level rewards.}, the model generates a response $y_0$ sampled from the policy $\pi(\cdot|x)$. Since $y_0$ is a token sequence generated autoregressively, the probability is defined as $\pi(y_0|x) = \prod_{i=1}^m \pi(y_{0,i}|x, y_{0,<i})$, where $y_0 = \{y_{0,1}, y_{0,2}, \dots, y_{0,m}\}$ denotes a sequence of $m$ tokens. Upon generation, a verifiable reward $r(x,y_0) \in \mathbb{R}$ quantifies the quality of the response, typically through deterministic rules such as unit tests, symbolic checkers, or exact string matching. The objective of RLVR is to optimize the policy $\pi$ to maximize the expected reward $\max_\pi \mathbb{E}_{y \sim \pi(\cdot|x)} [r(x,y)]$.




\textbf{Group Relative Policy Optimization (GRPO).} GRPO is a well-established, critic-free RLVR method. For prompt $x$, GRPO first samples a group of $n$ rollouts $\{y_1, y_2, \dots, y_n\}$ with corresponding rewards $\{r_1, r_2, \dots, r_n\}$, where each $y_i = \{y_{i,1}, y_{i,2}, \dots, y_{i,m}\}$. The objective is to minimize

\begin{equation}
\label{eq:grpo}
-\frac{1}{nM}\sum_{i=1}^n\sum_{j=1}^{|y_i|}\min\left[\frac{\pi_\theta(y_{i,j}|x,y_{i,<j})}{\pi_{\theta_\text{old}}(y_{i,j}|x,y_{i,<j})}A_{i}, \text{clip}\left(\frac{\pi_\theta(y_{i,j}|x,y_{i,<j})}{\pi_{\theta_\text{old}}(y_{i,j}|x,y_{i,<j})},1-\epsilon_1,1+\epsilon_2\right)A_{i}\right],
\end{equation}

where $\epsilon_2>\epsilon_1>0$ following DAPO~\cite{yu2025dapo}. $\pi_{\theta_{\text{old}}}$ represents the reference policy from the current sampling-training iteration, and $M$ is the upper limit of $y_i$'s length $|y_i|$.~\footnote{Following HPT~\cite{lv2025hpt} and Dr. GRPO~\cite{liu2025understanding}, we use the formulation without token mean.} The advantage $A_i$ is calculated relative to the group mean as $A_{i}=r(x,y_i)-\frac{1}{n}\sum_{k=1}^n r(x,y_k)$. Following recent work~\cite{lv2025hpt,ma2025relift}, we omit the KL regularizer and the standard deviation in advantage for simplicity.




\textbf{Direct Preference Optimization (DPO).} DPO~\cite{rafailov2023direct} is an offline RLHF framework that enables policy optimization directly from preference data without the need for an explicit reward model. Given an offline dataset $D$ consisting of triples $(x, y^+, y^-)$, where $y^+$ and $y^-$ represent the preferred and non-preferred responses respectively, DPO minimizes the following objective:

\begin{equation}
-\mathbb{E}_{(x,y^+,y^-)\sim D}\left[\log\sigma\left(\beta\log\frac{\pi_\theta(y^+|x)}{\pi_{\text{ref}}(y^+|x)}-\beta\log\frac{\pi_\theta(y^-|x)}{\pi_{\text{ref}}(y^-|x)}\right)\right],
\end{equation}

where $\pi_{\text{ref}}$ denotes the reference policy prior to training, and $\beta>0$ is a hyperparameter. 




\textbf{REINFORCE.} REINFORCE~\cite{Williams1992} is a fundamental policy gradient algorithm that remains widely utilized in contemporary RLVR research~\cite{ahmadian2024back}. In the RLVR setting, given a prompt $x$, a model response $y \sim \pi(\cdot|x)$, and a reward $r(x,y)$, REINFORCE aims to maximize the expected reward $\mathbb{E}_{x,y}[r(x,y)]$. The empirical loss gradient is defined as $-r(x,y) \cdot \nabla \log \pi(y|x)$.

\section{Methodology}
\label{sec:method}

In this section, we first delineate the three unique challenges inherent in few-shot demonstration-guided RLVR and identify the vital components of a post-training objective designed to address them (Sec.~\ref{sec:motivation}). We then introduce our algorithm, FEST, in Sec.~\ref{sec:fest}, followed by its variant, FEST-GRPO, to address its limitation on gradient mismatch in Sec.~\ref{sec:grpo}.


\subsection{The Three Unique Challenges and The Three Vital Components}
\label{sec:motivation}

In this framework, we optimize a Large Language Model (LLM) using two distinct datasets: a few-shot SFT dataset $D_E$ containing \textbf{E}xpert-curated reasoning traces, and a large-scale, answer-only (and thus \textbf{I}mperfect) RL dataset $D_I$. Our primary objective is to fully exploit the minimal reasoning traces in $D_E$ to enhance model performance beyond what is achievable through standard RLVR on $D_I$ alone.


This setting introduces three unique challenges: \textit{(i) No on-demand data:} Due to limited expert access, demonstrations cannot be generated on-demand for arbitrary questions where the model fails. This removes the flexibility assumed in many prior works such as HPT~\cite{lv2025hpt}, ReLIFT~\cite{ma2025relift}, LUFFY~\cite{yan2025luffy} and MIFO~\cite{yuan2025mitigating}. 
\textit{(ii) Limited Semantic Coverage:} The limited volume of SFT data is insufficient to cover the broad reasoning paradigms required. Furthermore, unlike specialized few-shot works like LIMOv2~\cite{ye2025limo}, we do not assume a carefully curated pipeline; $D_E$ may simply be a random batch of samples. 
\textit{(iii) Overfitting Risk:} Given the minuscule size of $D_E$, repeated training over multiple epochs risks severe overfitting, which can degrade the model's general reasoning capability.


We identify three vital components to address these challenges: \textit{\texthl[backgroundteal]{supervised learning}}, \textit{\texthl[backgroundorange]{on-policy learning}}, and \textit{\texthl[backgroundred]{adaptive weight scheduling}}. We argue that these components must be carefully integrated when training on $D_E$. First, \texthl[backgroundteal]{supervised learning} is essential as it provides the only source of external knowledge beyond the binary reward signals in RLVR. Second, \texthl[backgroundorange]{on-policy learning} addresses the first and second challenges by allowing the model to evaluate its own rollouts against SFT traces. This expands the learning basis for the limited questions in $D_E$~\cite{min2024imitate} and mitigates exposure bias~\cite{Bengio2015ScheduledSampling}. Finally, \texthl[backgroundred]{adaptive weight scheduling} is crucial for tackling the third challenge. We employ a \textit{decaying weight} strategy, ensuring the model prioritizes learning from $D_E$ in early stages while refraining from overfitting as the RLVR signal on $D_I$ becomes more dominant. Similar principles are observed in HPT~\cite{lv2025hpt}, where the SFT data ratio is reduced to $2\%$ toward the end of training (See Appendix~\ref{sec:complement}).

In conclusion, we require an algorithm for $D_E$ that incorporates both supervised and on-policy loss terms, governed by a decaying weight. As we discuss below, semi-online DPO~\cite{chen2024self,tan2024sopo} serves as an ideal framework to satisfy these requirements.


\subsection{FEST: FEw-ShoT Demonstration-Guided RLVR}
\label{sec:fest}

We define our training objective as follows. We optimize the parameters $\theta$ of the LLM policy $\pi_\theta$ using a semi-online DPO loss on the few-shot dataset $D_E$, where the SFT data serves as the preferred rollout $y^+$ and the RL-generated data acts as the non-preferred rollout $y^-$. As established, we utilize a GRPO loss for the answer-only dataset $D_I$ and a semi-online DPO loss for the few-shot dataset $D_E$. Specifically, we use $L = c\cdot L_E + L_I$ where
\begin{equation}
\label{eq:fest-main}
\begin{aligned}
    L_E&=-\mathbb{E}_{(x,y^+)\sim D_E,y^-\sim\pi_{\theta_{\text{old}}}(\cdot|x)}\left[\log\sigma\left(\beta\cdot r^+-\beta\cdot r^-\right)\right], \quad\text{and}\\
    L_I&=\mathbb{E}_{x\sim D_I,y\sim\pi_{\theta_\text{old}}(\cdot|x)}\frac{-1}{nM}\sum_{i=1}^n\sum_{j=1}^{|y_i|}\min\left[\frac{\pi_\theta(y_{i,j}|x,y_{i,<j})}{\pi_{\theta_\text{old}}(y_{i,j}|x,y_{i,<j})}A_{i},\right. \\
    &\qquad\left.
    \text{clip}\left(\frac{\pi_\theta(y_{i,j}|x,y_{i,<j})}{\pi_{\theta_\text{old}}(y_{i,j}|x,y_{i,<j})},1-\epsilon_1,1+\epsilon_2\right)A_{i}\right].\\
\end{aligned}
\end{equation}
In this objective, $r^+ = \log \frac{\pi_\theta(y^+|x)}{\pi_{\text{ref}}(y^+|x)}$, $r^- = \log \frac{\pi_\theta(y^-|x)}{\pi_{\text{ref}}(y^-|x)}$, $\sigma(x)=\frac{1}{1+e^{-x}}$ is the sigmoid function, and $c > 0$ is a constant coefficient. The detailed training pseudo-code is provided in Appendix~\ref{sec:pseudo}.


To justify the selection of the semi-online DPO loss for $L_E$, we examine its gradient:
\begin{equation}
\label{eq:grad}
\nabla_\theta L_E
=
-\beta\,
\mathbb{E}_{(x,y^+)\sim D_E,\, y^-\sim\pi_{\theta_\text{old}}}
\Big[
\mathhl[backgroundred]{\sigma(\beta (r^- - r^+))}
\cdot
\big(
\mathhl[backgroundteal]{\nabla \log \pi_\theta(y^+|x)}
-
\mathhl[backgroundorange]{\nabla \log \pi_\theta(y^-|x)}
\big)
\Big].
\end{equation}
The three terms in the gradient align precisely with our previously identified vital components: {\texthl[backgroundteal]{supervised learning}}, {\texthl[backgroundorange]{on-policy training}}, and {\texthl[backgroundred]{decaying weights}}. Theoretically, as demonstrated in SPIN~\cite{chen2024self}, this paradigm is equivalent to an adversarial training process. In each iteration, a discriminator optimizes a loss inspired by Integral Probability Metrics (IPM)~\cite{Muller1997} to differentiate $r^+$ and $r^-$, while the LLM policy acts as the generator with a closed-form solution. Further theoretical details are provided in Appendix~\ref{sec:adversarial}.


However, this standard DPO paradigm applies uniform learning strength across all data in $D_E$, failing to account for varying task difficulty. For simpler questions, deviations from the SFT demonstration should be tolerated, whereas the model should prioritize learning from SFT traces for tasks it cannot solve independently. For this, we apply an adaptive $\beta$ strategy based on model solvability. Specifically, for a batch of $n$ rollouts $\{y_1^-, \dots, y_n^-\}$ with binary rewards $r \in \{0, 1\}$, we define $\beta$ for the pair $(x, y_i^-)$ as:
\begin{equation}
    \beta(x,y_i^-) = 
\begin{cases}
\beta_1, & \text{if } \forall j\in\{1,2,\dots,n\}, r(x,y^-_j)=0 \\
\beta_2, & \text{if } r(x,y_i^-)=0 \text{ and }\exists\ j\in\{1,2,\dots,n\}, r(x,y_j^-)=1 \\
\beta_3, & \text{if } r(x,y_i^-)=1 \\
\end{cases}
\end{equation}
where $\beta_1, \beta_2, \beta_3$ are constants. This allows us to control the learning strength of different sources of data in a more fine-grained manner, differentiating unsolvable questions ($\beta_1$), RLVR-solvable questions ($\beta_2$) and correct rollouts ($\beta_3$).



\begin{remark}
While DPO is often criticized for its inability to effectively flip preferences~\cite{chen2024preference} and the dominance of the rejected response in the loss term~\cite{cho2025rethinking}, these characteristics are not detrimental in our setting. Our objective on $D_E$ is to ``regularize'' the model toward expert traces and catalyze RLVR performance—akin to an online version of TD3+BC~\cite{fujimoto2021minimalist}—rather than enforcing strict preference or total rejection of non-preferred samples, particularly as agent rollouts are often correct.
\end{remark}


\begin{remark}
The unique requirement of long-chain reasoning on few-shot data necessitates a distinct choice of $\beta$ (0.001--0.1) compared to standard DPO practices (0.1--0.2)~\cite{rafailov2023direct}. This is because the extended sequence lengths and repeated training on sparse data lead to significantly larger log-ratio differences. See Appendix~\ref{sec:ablation_hyperparam} for a full hyperparameter analysis.
\end{remark}


\subsection{FEST-GRPO: Mitigation of Gradient Mismatch}
\label{sec:grpo}

While the paradigm described in Sec.~\ref{sec:fest} is effective for few-shot demonstration-guided RLVR, a critical challenge persists: DPO for $L_E$ is a \textit{sequence-level} objective, where the probability inside the log-sigmoid represents the joint probability of the entire response. In contrast, GRPO for $L_I$ is a \textit{token-level} algorithm that applies clipping independently to each token. This structural mismatch often results in significant differences in gradient magnitudes (see Appendix~\ref{sec:complement}), necessitating exhaustive tuning of the coefficient $c$ to balance the gradients.


To mitigate this mismatch, we re-examine the {\texthl[backgroundred]{decaying weight}} and {\texthl[backgroundorange]{on-policy}} components of the gradient in Eq.~\eqref{eq:grad}: $\mathbb{E}_{x \sim D_E, y^- \sim \pi_{\theta_\text{old}}} \left[ \mathhl[backgroundred]{\beta \cdot \sigma(\beta(r^- - r^+))} \cdot \mathhl[backgroundorange]{\nabla \log \pi_\theta(y^-|x)} \right]$. By comparing this term to the REINFORCE gradient in Sec.~\ref{sec:prelim}, it becomes evident that this component is \textit{functionally equivalent to REINFORCE with a negative reward} defined as $-\beta \cdot \sigma(\beta(r^- - r^+)) < 0$. Similarly, the {\texthl[backgroundteal]{supervised learning}} component can be interpreted as weighted SFT with a positive weight $\beta \cdot \sigma(\beta(r^- - r^+)) > 0$. This observation provides a novel perspective on the DPO loss $L_E$:


\begin{center}
\textit{Semi-online DPO $\approx$ {\texthl[backgroundorange]{REINFORCE with negative reward}} + {\texthl[backgroundteal]{weighted SFT}}.} 
\end{center}

Under this interpretation, the solution to the gradient mismatch follows intuitively: we \textbf{substitute REINFORCE with GRPO}. We denote this variant as \textbf{FEST-GRPO}. This method retains the $L_I$ component in Eq.~\eqref{eq:fest-main} while replacing the DPO-based $L_E$ with a hybrid of weighted SFT and a GRPO loss applied to $D_E$.

\begin{remark}
While several recent works explore the decomposition of the DPO objective~\cite{wu2025takes,du2026rlhf,roux2025tapered}, our work establishes a formal equivalence between these specific algorithms, thereby extending the unified post-training framework proposed by HPT~\cite{lv2025hpt}. See Appendix~\ref{sec:extended_related} for detailed comparisons.
\end{remark}

\begin{remark}
The efficacy of RL with purely negative rewards is supported by prior theoretical work. Zhu et al.~\cite{zhu2025surprising} demonstrated that negative RL redistributes probability mass toward other feasible solutions, thereby preventing overfitting and facilitating more robust exploration.
\end{remark}





\section{Experiments}
\label{sec:exp}

In this section, we evaluate FEST across various settings and benchmarks to investigate the following research questions: 
(i) Can both the DPO and GRPO variants of FEST enhance RLVR performance using demonstrations as few as possible? (Sec.~\ref{sec:main}); 
(ii) How does FEST scale with the number of shots? (Sec.~\ref{sec:scale}); 
(iii) To what extent is the performance sensitive to the choice of datasets? (Sec.~\ref{sec:choice}); and 
(iv) What are the individual contributions of each component, and can FEST generalize to Out-Of-Distribution (OOD) test sets? (Sec.~\ref{sec:ablation}).


\textbf{Training Recipe.} We fine-tune the Qwen2.5-Math-1.5B~\cite{yang2024qwen25} model for 600 steps on two NVIDIA GH200 (96GB) GPUs. Following prior work~\cite{yan2025luffy,lv2025hpt,ma2025relift}, we utilize the OpenR1-Math-46K-8192 dataset as our primary source. We randomly sample 128 problems with expert reasoning traces to form $D_E$, while the remaining data serve as the answer-only dataset $D_I$ for reward verification. The number 128 is the batch size from prior works~\cite{lv2025hpt,ma2025relift}, which means an epoch on the dataset can be fitted into a single step. We generate $n=8$ rollouts per prompt with a temperature of $1.0$ and a maximum sequence length of 8192. We employ the AdamW optimizer~\cite{loshchilov2017decoupled} with a cosine learning rate decay from $1\times10^{-5}$ to $5\times10^{-6}$. The training uses a global batch size of 128 questions from each of $D_E$ and $D_I$, and a mini-batch size of 512 rollouts (all GPUs combined).


\textbf{Baselines.} We compare FEST against the following baselines: 
(i) \textit{Vanilla RLVR} with pure GRPO; 
(ii) \textit{Multi-objective approaches}, such as SRFT~\cite{fu2025srft}, LUFFY~\cite{yan2025luffy}, and CHORD-$\phi$~\cite{zhang2025onpolicy}; and 
(iii) \textit{RL-SFT switching strategies}, including MIFO~\cite{yuan2025mitigating}, HPT~\cite{lv2025hpt}, and ReLIFT~\cite{ma2025relift}. 
We omit several related methods for specific reasons: SuperRL~\cite{liu2025superrl} and DyME~\cite{koksal2025few} are functionally identical to HPT in this context, while SASR~\cite{chen2025sasr} focuses on simpler tasks and lacks an accessible codebase with readme files. Pure SFT and SPIN~\cite{chen2024self} were excluded after failing to achieve competitive results (see Appendix~\ref{sec:sftspin}). Most baseline results were obtained with our own implementations, with two exceptions: SRFT and MIFO. SRFT is not open-source and does not work with the implementation in HPT codebase, thus we test their official checkpoint; MIFO does not publish code or checkpoint, and we directly take the results from their paper. See Appendix~\ref{sec:extended_related} for an introduction to the baselines.


\textbf{Evaluation and Metrics.} Following the evaluation protocol in HPT~\cite{lv2025hpt}, we assess performance on six prominent mathematical reasoning benchmarks: AIME25~\cite{li2024numinamath} (30 questions), AMC23~\cite{li2024numinamath} (40 questions), AIME24~\cite{li2024numinamath} (30 questions), MATH-500~\cite{hendrycks2021measuring} (500 questions), OlympiadBench~\cite{he2024olympiadbench} (674 questions), and Minerva~\cite{lewkowycz2022solving} (272 questions). To ensure statistical stability, particularly on smaller benchmarks, we report the mean and standard deviation across 8 rollouts (Avg@8). We also report Pass@8 (the percentage of questions with at least one correct response in 8 trials) to demonstrate the model's potential for further RL-driven improvement. Following ReLIFT~\cite{ma2025relift}, we report the result at 600 steps.


\subsection{Main Results}
\label{sec:main}

\begin{table}[ht]
    \scriptsize
    \centering
    \caption{Average test set accuracy using a 128-shot configuration (higher is better). The suffix ``-G'' denotes RL training also performed on the few-shot \textbf{G}old dataset ($D_E$). SRFT and MIFO utilize the \textbf{full dataset}, while all other methods are evaluated under the same 128-shot constraint. Results demonstrate that FEST variants not only achieve superior performance, but also represent the only methodology to yield significant gains over the vanilla RL baseline in this sparse-data regime.}
    \begin{tabular}{lccccccc}
    \toprule
        \multicolumn{1}{c}{Methods} & AIME25 & AMC23 & AIME24 & MATH-500 & Olympiad & Minerva  & Average \\
        \midrule 
       SRFT$^*$ (ICLR`26) & 7.91$\pm$2.86 & 51.25$\pm$4.15 & 10.83$\pm$3.23 & 72.6$\pm$1.47 & 38.58$\pm$1.20 & 29.09$\pm$1.15 & 35.05$\pm$1.04\\
       MIFO$^*$ (Preprint) &  12.0 & N/A  & 19.2 & 78.8 & 43.3 & N/A & N/A \\
       RL & 11.67$\pm$4.41 & 54.37$\pm$5.41 & 17.91$\pm$3.31 & 77.78$\pm$1.24 & 43.25$\pm$1.27 & 33.78$\pm$1.49 & 39.79$\pm$1.04\\ 
       RL-G & 14.17$\pm$4.33 & 57.5$\pm$5.15 & 15.83$\pm$3.23 & 78$\pm$1.27 & 44.49$\pm$0.86 & 33.36$\pm$1.32 & 40.55$\pm$1.03 \\
       LUFFY (NeurIPS`25) & 9.58$\pm$3.87 & 55$\pm$1.77 & 14.17$\pm$3.23 & 74.88$\pm$1.17 & 43.49$\pm$0.69 & 30.28$\pm$5.50 & 37.90$\pm$0.50 \\
       CHORD-$\phi$ (ICLR`26) & 8.75$\pm$4.06 & 53.13$\pm$4.64 & 13.33$\pm$2.89 & 76.03$\pm$0.89 & 42.47$\pm$0.75 & 31.76$\pm$1.55 & 37.56$\pm$1.47 \\
       HPT (Preprint) & 10.83$\pm$2.76 & 55$\pm$4.33 & 13.75$\pm$3.09 & 76.65$\pm$0.70 & 43.14$\pm$0.92 & 33.13$\pm$2.03 & 38.75$\pm$1.19\\
       HPT-G & 3.33$\pm$3.33 & 46.88$\pm$1.65 & 10.41$\pm$1.99 & 66.7$\pm$1.10 & 34.34$\pm$1.31 & 30.51$\pm$1.23 & 32.02$\pm$0.81\\
       ReLIFT (ICLR`26) & 11.25$\pm$3.30 & 57.81$\pm$2.32 & 15.83$\pm$2.20 & 79.03$\pm$8.69 & 44.35$\pm$7.94 & 34.79$\pm$1.27 & 40.51$\pm$1.03\\
       ReLIFT-G & 9.16$\pm$3.63 & 54.06$\pm$3.29 & 14.58$\pm$2.32 & 76.6$\pm$1.16 & 42.28$\pm$0.73 & 32.44$\pm$0.86 & 38.19$\pm$0.77 \\
       
       \midrule
       FEST-DPO & 11.67$\pm$4.41 & \textbf{59.38$\pm$3.25} & \textbf{20.41$\pm$3.51} & 80.48$\pm$1.13 & 46.80$\pm$1.01 & 33.18$\pm$1.85 & 41.98$\pm$1.24\\
       FEST-GRPO & \textbf{14.58$\pm$3.70} & 57.81$\pm$5.36 & 18.33$\pm$1.67 & \textbf{81.1$\pm$1.27} & \textbf{47.00$\pm$0.95} & \textbf{35.45$\pm$1.81} & \textbf{42.36$\pm$1.26} \\
    \bottomrule
    \end{tabular}
    
    \label{tab:main}
\end{table}

Tab.~\ref{tab:main} presents the primary results of this study, demonstrating that our proposed method outperforms all established baselines. We highlight three key observations: (i) \textbf{Instability in Naive SFT and RL Gradient Integration.} In Tab.~\ref{tab:main}, we evaluate pure RL, ReLIFT, and HPT appended with a ``-G'' suffix, indicating that RL was also conducted on the ``Gold'' few-shot dataset $D_E$. Surprisingly, both HPT-G and ReLIFT-G perform significantly worse than their variants that utilize only SFT on $D_E$. An investigation of the training curves reveals that both models suffer from abrupt performance drops on the training set mid-process (see Appendix~\ref{sec:withgold}). We hypothesize that this instability arises from rapid distribution shifts induced by SFT on a dataset already heavily overfitted by RL, reinforcing the necessity of our decaying weight strategy. (ii) \textbf{Efficacy of the Pure RL Baseline.} Contrary to findings in prior work~\cite{lv2025hpt,ma2025relift,yan2025luffy}, we observe that pure RL remains a formidable baseline when the learning rate is optimized (specifically, increased from 1e-6 to 5e-6\footnote{We also evaluated pure RL using our specific learning rate schedule, which did not yield further improvements.}). Under this configuration, pure RL achieves performance parity with ReLIFT on the full dataset.
(iii) \textbf{Consistency in Outperforming RL-G Variants.} FEST is the only method that consistently surpasses both pure RL and RL-G. While RL-G may achieve high nominal accuracy, it suffers from severe overfitting on the limited $D_E$ dataset, resulting in a significantly lower Pass@8 (see Tab.~\ref{tab:pass8}). This reduction highlights a diminished potential for further improvement---a pitfall FEST avoids by maintaining high exploration capability.

\begin{table}[ht]
    \scriptsize
    \centering
    \caption{Pass@8 performance representing the models' upper-bound exploration potential. Notably, while RL-G achieves relatively high average accuracy, its significantly lower Pass@8 indicates limited reasoning diversity and a lower ceiling for further performance refinement. In contrast, FEST variants consistently maintain a superior Pass@8, demonstrating robust potential for subsequent optimization.}
    \begin{tabular}{ccccccccc}
    \toprule
        Methods &  AIME25 & AMC23 & AIME24 & MATH-500 & Olympiad & Minerva  & Average \\
        \midrule 
       SRFT$^*$ & 23.33 & 72.5 & 30 & 91 & 63.54 & 48.90 & 54.87\\
       RL & \textbf{36.67} & 85 & 33.33 & 91.4 & 63.1 & 48.52 & 59.67 \\ 
       RL-G & 33.33 & 72.5 & 23.33 & 90.6 & 63.69 & 45.59 & 54.84 \\
       LUFFY  & 30 & 82.5 & 26.67 & 89.6 & 61.31 & 45.22 & 55.88 \\
       HPT & 26.67 & \textbf{87.5} & 26.67 & 91.2 & 64.73 & 51.84 & 58.10 \\
       HPT-G  & 16.67 & 70 & 16.67 & 80.6 & 53.57 & 45.22 & 47.12 \\
       ReLIFT & 20 & 85 & 30 & 91.2 & 64.88 & 49.26 & 56.72 \\
       ReLIFT-G  & 26.67 & 77.5 & 30 & 89.8 & 61.76 & 47.43 & 55.52\\
       CHORD-$\phi$ & 20 & 80 & 26.67 & 91.6 & 62.5 & 50.37 & 55.19 \\
       \midrule
       FEST-DPO &  26.67 & 82.5 & \textbf{40} & \textbf{92.8} & 65.17 & \textbf{53.31} & 60.08 \\
       FEST-GRPO & \textbf{36.67} & \textbf{87.5} & 33.33 & 92 & \textbf{65.77} & 51.1 & \textbf{61.06}\\
    \bottomrule
    \end{tabular}
    
    \label{tab:pass8}
\end{table}

\begin{wrapfigure}[19]{r}{0.45\linewidth}
    \centering
    \includegraphics[width=\linewidth]{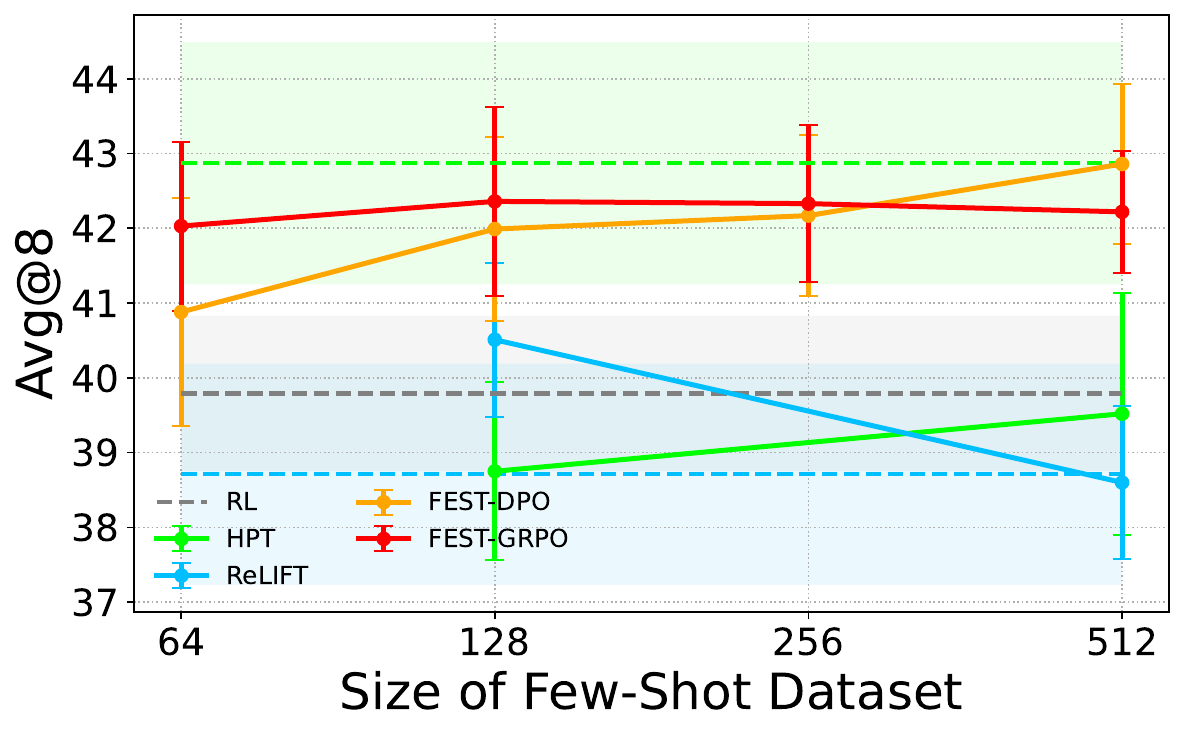}
    \caption{Performance scalability across varying shot counts. Dashed lines represent baseline results utilizing the full 46K SFT dataset. While FEST-GRPO provides higher robustness in the extreme few-shot case (64 shots), FEST-DPO exhibits stronger scaling ability with more data.}
    \label{fig:scaling}
\end{wrapfigure}

\subsection{Scaling with Shot Counts}
\label{sec:scale}
To evaluate the scalability of our approach across varying sizes of $D_E$, we further test our method with 64, 256, and 512 examples. The results, illustrated in Fig.~\ref{fig:scaling}, demonstrate that our method can work even with as few as 64 shots. Notably, while FEST-GRPO exhibits superior stability in extreme low-data regimes, FEST-DPO demonstrates more favorable scaling properties, eventually achieving performance comparable to HPT trained on the full 46K SFT dataset.



\subsection{Consistency of Performance Gain Across Different \texorpdfstring{$D_E$}{DE}}
\label{sec:choice}

To evaluate the robustness of our algorithm across varying $D_E$, we test FEST on several alternative few-shot datasets ($D_E$): (i) two additional random 128-shot splits from OpenR1; and (ii) LIMOv2-8192, a subset of 257 examples from LIMOv2~\cite{ye2025limo} with Chain-of-Thought (CoT) traces under 8,192 tokens.\footnote{Due to computational constraints, we exclude LIMOv2 examples with longer reasoning traces.} Tab.~\ref{tab:limov2} summarizes the results, demonstrating that FEST consistently achieves performance gains across different $D_E$. See further training details regarding LIMOv2 in Appendix~\ref{sec:curves}.


\begin{table}[ht]
    \centering
    \scriptsize
    \caption{Avg@8 performance of FEST across different $D_E$ settings, with the results from Sec.~\ref{sec:main} denoted as ``OpenR1 (dataset 1)''. The upper and middle blocks correspond to FEST-DPO and FEST-GRPO, respectively, while the lower block shows RL result for reference. The results indicate that our framework is consistently capable of extracting significant performance improvements from different few-shot demonstrations.}
    \begin{tabular}{lccccccc}
    \toprule
         & AIME25 & AMC23 & AIME24 & MATH-500 & Olympiad & Minerva & Average \\
         \midrule
         \multicolumn{8}{c}{\textbf{FEST-DPO}} \\
    \midrule
       LIMOv2-8192 & 14.16$\pm$3.23 & 54.69$\pm$2.63 & 17.5$\pm$5.71 & 80.23$\pm$1.21 & 46.28$\pm$0.71 & 34.37$\pm$1.34 & 41.21$\pm$1.25 \\
       OpenR1 (dataset 1) & 11.67$\pm$4.41 & 59.38$\pm$3.25 & 20.41$\pm$3.51 & 80.48$\pm$1.13 & 46.80$\pm$1.01 & 33.18$\pm$1.85 & 41.98$\pm$1.24\\
       OpenR1 (dataset 2) & 14.58$\pm$6.44 & 53.75$\pm$4.67 & 17.08$\pm$3.09 & 79.7$\pm$0.76 & 46.56$\pm$0.57 & 36.02$\pm$1.09 & 41.28$\pm$1.01 \\
       OpenR1 (dataset 3) & 12.91$\pm$2.60 & 56.88$\pm$2.07 & 18.33$\pm$2.89 & 80.6$\pm$0.75 & 46.24$\pm$1.20 & 35.47$\pm$1.25 & 41.74$\pm$0.81 \\
    \midrule
    \multicolumn{8}{c}{\textbf{FEST-GRPO}} \\
    \midrule
    LIMOv2-8192 & 14.17$\pm$3.23 &55$\pm$6.25 & 20.83$\pm$2.76 & 79.55$\pm$0.79 & 46.11$\pm$0.90 & 34.15$\pm$1.99 & 41.63$\pm$1.16 \\
    OpenR1 (dataset 1) & 14.58$\pm$3.70 & 57.81$\pm$5.36 & 18.33$\pm$1.67 & 81.1$\pm$1.27 & 47.00$\pm$0.95 & 35.45$\pm$1.81 & 42.38$\pm$1.26 \\
       OpenR1 (dataset 2) & 12.08$\pm$3.70 & 58.75$\pm$4.51 & 15.83$\pm$2.20 & 80.58$\pm$1.10 & 47.63$\pm$0.73 & 33.08$\pm$1.50 & 41.33$\pm$1.46 \\
       OpenR1 (dataset 3) & 12.08$\pm$5.76 & 58.13$\pm$4.28 & 16.67$\pm$3.33 & 80$\pm$1.24 & 46.88$\pm$1.19 & 33.36$\pm$1.18 & 41.18$\pm$1.68 \\
    \midrule
    RL (no $D_E$) & 11.67$\pm$4.41 & 54.37$\pm$5.41 & 17.91$\pm$3.31 & 77.78$\pm$1.24 & 43.25$\pm$1.27 & 33.78$\pm$1.49 & 39.79$\pm$1.04\\
    \bottomrule
    \end{tabular}
    
    \label{tab:limov2}
\end{table}

\subsection{Ablations}
\label{sec:ablation}

Due to the page limit, the details of ablation for hyperparameter $\beta$ is deferred to Appendix~\ref{sec:ablation_hyperparam}. Generally, we find FEST is reasonably robust to the choice of $\beta$, and works best with $\beta\in[0.001,0.1]$.


\subsubsection{Components}

In Sec.~\ref{sec:motivation}, we identified three vital components of few-shot demonstration-guided RLVR. To quantify the contribution of each component to the final performance, we conduct an ablation study under the experimental configuration described in Sec.~\ref{sec:main}. The results, summarized in Tab.~\ref{tab:ablation}, demonstrate that optimal performance is achieved only through the synergy of all three components. Specifically, the poor performance of the ``RL-G with our weight'' variant stems from model divergence: when the model is repeatedly exposed only to negative signals on $D_E$ during late-stage training, it lacks sufficient constructive guidance to maintain a coherent policy, eventually leading to model collapse.


\begin{table}[ht]
    \small
    \centering
    \caption{Ablation study of the vital components identified in Sec.~\ref{sec:motivation}. The results indicate that peak performance relies on the effective collaboration of all three components.}
    \begin{tabular}{ccccccccc}
    \toprule
       {\texthl[backgroundteal]{Supervised}} & {\texthl[backgroundorange]{On-Policy}} & {\texthl[backgroundred]{Decaying Weight}} & Equivalent to & Avg@8 \\
    \midrule
      \redcross & \redcross & \redcross / \greencheck &  RL & 39.79$\pm$1.04 \\
      \redcross & \greencheck & \redcross & RL-G &  40.55$\pm$1.03 \\
      \redcross & \greencheck & \greencheck & RL-G with our weight &  28.83$\pm$0.92 \\
      \greencheck & \redcross / \greencheck & \redcross & Fixed weight SFT+RL on \textbf{full dataset} &  37.26$\pm$1.12 \\
       \greencheck & \redcross & \greencheck & RL + few-shot SFT with our weight & 39.91$\pm$0.67 \\
       \greencheck & \greencheck & \greencheck & FEST-GRPO & \textbf{42.38$\pm$1.26} \\
    \bottomrule
    \end{tabular}
    
    \label{tab:ablation}
\end{table}

\subsubsection{Out-of-Distribution Dataset}

To further evaluate the cross-domain generalization of our approach, we follow the evaluation protocol of ReLIFT~\cite{ma2025relift}, and assess the model trained in Sec.~\ref{sec:main} on the MMLU-Pro benchmark~\cite{wang2024mmlu}, which consists of over 12,000 problems. The Pass@1 results are presented in Tab.~\ref{tab:mmlu}. Our results indicate that FEST generalizes robustly to OOD tasks, achieving superior performance compared to all evaluated baselines under this zero-shot setting.


\begin{table}[ht]
    \centering
    \scriptsize
    \caption{Pass@1 performance on the OOD MMLU-Pro benchmark. FEST variants consistently achieve the highest results, highlighting strong generalization capabilities beyond  training data.}
    \begin{tabular}{cccccccccc}
    \toprule
        FEST-DPO (ours) & FEST-GRPO (ours) & RL & RL-G & LUFFY & HPT & ReLIFT & SRFT & CHORD-$\phi$ & Random \\
    \midrule
        \textbf{38.68} & 36.32 & 34.81 & 33.18 & 31.72 & 32.83 & 33.99 & 33.54 & 35.82 & 10 \\
    \bottomrule
    \end{tabular}
    
    \label{tab:mmlu}
\end{table}

\section{Related Work}

\textbf{Demonstration-Guided RLVR.} A fundamental limitation of RLVR is its difficulty in surpassing the inherent capabilities of the base model~\cite{zhao2025echo,cheng2026reasoning,yue2025does}. While RL effectively sharpens the output distribution toward correct answers~\cite{wu2025invisible}, it often fails to explore solutions beyond the model's initial reasoning capacity. To address this, various hybrid post-training methods integrating SFT with RL have been proposed~\cite{jiang2026supervised}, leveraging expert demonstrations to provide guidance beyond the base model's limits. The most prevalent approaches are reward-based, conducting SFT on problems where the model fails to receive a positive reward~\cite{lv2025hpt,ma2025relift,liu2025superrl,liu2025empowering,yuan2025mitigating}. Other integration strategies include using fixed data ratios~\cite{yan2025luffy}, gradient-norm balancing~\cite{chen2025sasr}, token probability weighting~\cite{zhang2025onpolicy,yuan2025mitigating}, entropy-based regularization~\cite{fu2025srft}, and trajectory blending~\cite{liu2025uft,huang2025blending}. However, as shown in Tab.~\ref{tab:teaser}, these methods typically require substantial SFT data, which is often prohibitively expensive to acquire~\cite{wsj_mercor_2026}. In contrast, FEST is specifically designed to enhance RLVR performance using only a few-shot SFT dataset.


\textbf{Few-Shot LLM Post-Training.} In response to the tension between the increasing demand for high-quality reasoning data and its high cost, the community has actively explored post-training LLMs with minimal data via SFT~\cite{muennighoff2025s1}, DPO~\cite{singh2025fspo}, or RL~\cite{zhu2025drive}. Existing few-shot SFT and DPO research generally falls into two categories: (i) \textit{careful curation} via heuristic metrics—such as CoT length~\cite{wei2022chain}, diversity, and difficulty—exemplified by s1K~\cite{muennighoff2025s1} and LIMO~\cite{ye2025limo}; and (ii) \textit{automated data selection} based on distribution similarity to the base model~\cite{zhang2025best}, gradient-based performance estimation~\cite{wang2025nice}, embedding-based diversity~\cite{bukharin2024data,yang2025measuring,cao2023instruction} or regression of multiple factors~\cite{cao2023instruction}. On the RL side, researchers have investigated few-shot RLVR~\cite{li2025limr,fang2025serl,li2025confidence,koksal2025few} using question-only data~\cite{yang2025trapo} or extremely limited question sets~\cite{wang2025reinforcement}. However, none of these works focus on combining RLVR with few-shot SFT data as we do. Furthermore, unlike many prior works, FEST does not require labor-intensive dataset selection or curation to be effective.


\textbf{(Semi-)Online DPO and Self-Play Preference Learning.} To enable models to learn from self-generated data and mitigate exposure bias~\cite{Bengio2015ScheduledSampling}, several studies have explored iterative~\cite{xu2023some,min2024imitate,wang2026triplets} or online DPO~\cite{guo2024direct,xiong2023iterative,bai2025online}, where the preference dataset is updated during training using the agent's own rollouts. FEST adopts a similar principle but operates as a \textit{semi-online} DPO algorithm, where only the non-preferred responses are generated on-the-fly~\cite{pan2025matters}. Under this definition, the closest works to ours are SoPo~\cite{tan2024sopo} and SPIN~\cite{chen2024self}, both of which utilize static SFT data as preferred examples and agent rollouts as non-preferred ones. However, SoPo is specifically tailored for human motion generation with diffusion models. While SPIN demonstrates that this practice is equivalent to adversarial training via an Integral Probability Metric (IPM)~\cite{Muller1997} objective, FEST differs in several critical ways: (i) while SPIN targets RLHF with large SFT datasets, FEST is optimized for few-shot demonstration-guided RLVR; (ii) unlike the iterative nature of SPIN or STILL-2~\cite{min2024imitate}, FEST utilizes a truly online sampling paradigm\footnote{The verl implementation of SPIN (\url{https://verl.readthedocs.io/en/latest/algo/spin.html}) is online but deviates significantly from the original framework by omitting SFT data.}; (iii) inspired by RLVR research~\cite{lv2025hpt}, we introduce adaptive $\beta$ values to handle varying task difficulties and long-horizon reasoning—factors not considered in the SPIN framework; and (iv) we propose the FEST-GRPO variant, which leverages a token-level clipped gradient that is fundamentally different from the sequence-level SPIN objective.

\section{Conclusion}
\label{sec:conclusion}

In this work, we introduced a novel post-training paradigm, few-shot demonstration-guided RLVR, alongside a dedicated training framework, \textbf{FEST}. By elucidating the core challenges of training with sparse expert data, we designed a framework based on three vital components: (i) {\texthl[backgroundteal]{supervised learning}} from SFT data to provide expert guidance; (ii) {\texthl[backgroundorange]{on-policy learning}} with negative advantages to expand the exploration base and ensure robustness; and (iii) a {\texthl[backgroundred]{decaying weight}} strategy to prevent overfitting. Furthermore, by decomposing the DPO objective, we proposed \textbf{FEST-GRPO} to bridge the gradient magnitude gap between token-level GRPO and sequence-level DPO. Extensive experiments across multiple benchmarks demonstrate that FEST effectively boosts RLVR performance using as few as 128 randomly sampled demonstrations. We believe this paradigm and its associated algorithms offer a scalable solution to the growing scarcity of high-quality SFT data in the LLM community.


\textbf{Limitations and Future Work.} Due to computational resource constraints, our evaluation was mostly conducted with a 1.5B parameter model focused on math reasoning. Investigating how FEST scales to larger model architectures and generalizes to broader task domains, such as code generation and general instruction following, remains a significant and promising direction for future research.

\bibliography{ref}
\bibliographystyle{neurips_from2024}

\newpage
\appendix

\section*{Appendix for Boosting Reinforcement Learning with Verifiable Rewards via Randomly Selected Few-Shot Guidance}

This appendix is organized as follows. First, in Sec.~\ref{sec:impact}, we discuss the broader societal impact of our work. We then discuss more related works in Sec.~\ref{sec:extended_related}, provide implementation details of our method and baselines in Sec.~\ref{sec:implement}, and show more experimental results in Sec.~\ref{sec:more_results}. Finally, we report our computational resource in Sec.~\ref{sec:resources} and licenses in Sec.~\ref{sec:license}.

\section{Broader Impacts}
\label{sec:impact}

This work introduces a resource-efficient paradigm for post-training LLM agents using minimal expert data. Beyond its technical contributions, our research carries several broader societal implications:

\textbf{Ethical Data Sourcing and Labor Practices.} A significant portion of current LLM development relies on large-scale crowdsourcing for data annotation, which often involves ``ghost work'' that can lead to the economic and psychological exploitation of workers in the Global South. By demonstrating that performance can be significantly boosted with as few as 128 random samples, FEST reduces the systemic dependence on massive human-labeled datasets. This shift helps mitigate the ethical risks associated with labor exploitation and improves the sustainability of high-performance AI development.

\textbf{Democratization and Legal Compliance.} Current distillation practices, while effective, often inhabit a legal ``gray zone'' regarding intellectual property and proprietary API usage. Furthermore, the high cost of acquiring extensive SFT data creates a barrier to entry for smaller research entities. Our few-shot approach provides a compliant and cost-effective alternative, democratizing access to state-of-the-art RLVR techniques for academic and independent researchers who lack industrial-scale resources.

\textbf{Potential Risks and Mitigations.} Despite these benefits, we acknowledge that enhancing the reasoning capabilities of LLMs carries inherent risks. Improved mathematical and logical proficiency could be dual-purposed for malicious activities or accelerate technological displacement in certain labor markets. Furthermore, while RLVR targets ``verifiable'' rewards, there remains a risk of models developing ``reward-hacking'' behaviors or unintended biases if the verifiers are not perfectly specified. We believe that addressing these challenges requires a multi-stakeholder approach involving proactive safety alignment, robust policy frameworks, and transparent reporting from the research community.


\section{Extended Related Works}
\label{sec:extended_related}

\subsection{More Related Works}

\textbf{Unifying RL and SFT in LLM Post-Training.} The conventional paradigm for modern LLM post-training typically follows a decoupled, two-stage process: an initial ``cold start'' via SFT followed by RLVR to enhance reasoning capabilities and mitigate exposure bias~\cite{Guo2025DeepSeek}. While several state-of-the-art models employ multiple alternating stages of SFT and RL to handle diverse tasks---such as distinguishing between improving core reasoning ability and integration of Mixture-of-Experts (MoE)~\cite{shazeer2017outrageously} routing~\cite{yang2025qwen3,Guo2025DeepSeek}---the objectives of these stages remain largely distinct. However, recent findings have challenged this sequential approach, suggesting it may not consistently outperform pure RL due to its tendency to disrupt established behavioral patterns or induce overfitting~\cite{zhang2025nemotron,chen2025sft,zhang2025onpolicy}. Consequently, a burgeoning line of research has shifted toward a unified framework where SFT and RL are conducted jointly~\cite{yan2025luffy,lv2025hpt,ma2025relift,zhang2025onpolicy,fu2025srft}. Our work aligns with this movement but distinguishes itself by significantly reducing the dependency on extensive SFT datasets.


\textbf{DPO Gradient Decomposition.} Since the introduction of DPO, numerous studies have identified the underlying dual-nature of its objective: a combination of a weighted SFT loss and a negative reinforcement learning signal~\cite{wu2025takes,roux2025tapered,du2026rlhf}. This gradient decomposition has become a cornerstone for analyzing and refining DPO, leading to innovations such as the elimination of explicit preference pairs~\cite{ethayarajh2024kto}, the introduction of ``positive anchoring'' to ensure the absolute log-probability of preferred responses increases~\cite{pal2024smaug,d-oosterlinck-etal-2025-anchored,feng2024towards,guo2025proximalized,ma2025gradient,afanasyev2026slime}, more targeted safety alignment~\cite{zhao2025improving}, and the development of broader unified RLHF frameworks~\cite{du2026rlhf,he2026dual}. 
Despite these advancements, most existing works treat this structural decomposition as a conceptual analogy to general RL rather than establishing a formal mathematical equivalence between online DPO and specific algorithms like REINFORCE~\cite{Williams1992}. The closest efforts in this direction include TOPR~\cite{roux2025tapered}, which discusses off-policy REINFORCE, and 2-GRPO~\cite{wu2025takes}, which establishes equivalence between GRPO with two rollouts and online DPO. However, these studies focus on algorithmic variants that remain distinct from the semi-online DPO formulation proposed in this work.


\subsection{Semi-Online DPO and Adversarial Training}
\label{sec:adversarial}


\textbf{Preliminary: Integral Probability Metrics (IPM).} Integral Probability Metrics (IPMs)~\cite{Muller1997} provide a general framework for measuring the statistical distance between two probability distributions. Formally, given two distributions $p$ and $q$, the IPM is defined as:

\begin{equation}
\label{eq:ipm_intro}
\sup_{f\in\mathcal{F}}\left|\mathbb{E}_{x\sim P}[f(x)]-\mathbb{E}_{x\sim q}[f(x)]\right|,
\end{equation}

where $\mathcal{F}$ denotes a class of real-valued bounded measurable functions. The choice of $\mathcal{F}$ determines the metric; for instance, if $\mathcal{F}$ is the set of 1-Lipschitz functions, the IPM recovers the 1-Wasserstein distance.


Following the principles of Wasserstein inverse RL~\cite{WDAIL}, we consider an adversarial training process where a generator (the policy $\pi$) and a discriminator (the reward function $f$) engage in a minimax game. The discriminator seeks to maximize the discrepancy between the agent's policy $\pi$ and the expert behavioral policy $\pi^E$ from $D_E$. This yields the following objective:


\begin{equation}
\label{eq:ipm}
    \max_f\min_\pi \mathbb{E}_{x\sim D_E}\left[\mathbb{E}_{y\sim\pi_E(\cdot|x)}[f(x,y)]-\mathbb{E}_{y'\sim \pi(\cdot|x)}[f(x,y')]\right],
\end{equation}

where the absolute value is omitted assuming $\mathcal{F}$ is symmetric (i.e., $f \in \mathcal{F} \implies -f \in \mathcal{F}$). However, for any given $\pi$, this linear objective is unbounded for unconstrained $f$, which leads to potential instability during training. Therefore, we introduce a non-linear activation $g(\cdot)$. Specifically, we let $g(x) = -\log(1+e^{-x})$ be the log-sigmoid function (equivalent to the ``logistic loss'' in SPIN~\cite{chen2024self}), leading to:


\begin{equation}
\label{eq:spin}
\max_f\min_\pi \mathbb{E}_{x\sim D_E} g\left(\mathbb{E}_{y\sim\pi^E}[f(x,y)]-\mathbb{E}_{y'\sim \pi(\cdot|x)}[f(x,y')]\right),
\end{equation}

To make the inner minimization tractable, we assume the generator maximizes expectation of $f$ subject to a KL-regularization term anchored at the reference policy $\pi_{\text{ref}}$:


\begin{equation}
\label{eq:reg}
    \mathbb{E}_{x\sim D_E}\left[\mathbb{E}_{y'\sim\pi(\cdot|x)}[f(x,y')]-\beta\text{KL}(\pi\|\pi_{\text{ref}})\right]
\end{equation}

The analytical solution for the optimal policy is given by the Gibbs distribution: $\pi(y'|x) \propto \pi_{\text{ref}}(y'|x) \exp\left( \frac{f(x,y')}{\beta} \right)$. Solving for $f$, we obtain $f(x,y') = \beta \log \frac{\pi(y'|x)}{\pi_{\text{ref}}(y'|x)} + c(x)$. Substituting this back into Eq.~\eqref{eq:spin} results in:


\begin{equation}
\label{eq:beforejensen}
\min_\pi -\mathbb{E}_{x\sim D_E}\log\sigma(\beta\mathbb{E}_{y\sim\pi^E}[\log\frac{\pi(y|x)}{\pi_{\text{ref}}(y|x)}]-\beta\mathbb{E}_{y'\sim\pi}[\log\frac{\pi(y'|x)}{\pi_{\text{ref}}(y'|x)}]),
\end{equation}

Finally, by invoking Jensen's inequality~\cite{jensen1906convex} and the concavity of the log-sigmoid function, we derive a tractable upper bound for this objective:


\begin{equation}
\label{eq:afterjensen}
\min_\pi -\mathbb{E}_{(x,y)\sim D_E,y'\sim\pi(\cdot|x)}\log\sigma(\beta\log\frac{\pi(y|x)}{\pi_{\text{ref}}(y|x)}-\beta\log\frac{\pi(y'|x)}{\pi_{\text{ref}}(y'|x)}),
\end{equation}

which corresponds precisely to our semi-online DPO loss.


\begin{remark}
Our derivation is slightly different from SPIN's original framework, where they see $\mathbb{E}_{(x,y)\sim D^E, y'\sim\pi(\cdot|x)}\left[g(f(x,y)-f(x,y'))\right]$ as a generalization of Eq.~\eqref{eq:ipm}, thus skipping the Jensen's inequality from Eq.~\eqref{eq:beforejensen} to Eq.~\eqref{eq:afterjensen}. Our derivation further proves the numerical connection by bounds between the ``generalized loss'' and the original IPM loss, and also does not make assumptions on function class as in their Eq. (4.5).
\end{remark}

\subsection{Unifying Post-Training Framework}
\label{sec:hpt}
HPT~\cite{lv2025hpt} establishes a unified post-training framework that synthesizes the gradients of various algorithms into a generalized policy gradient estimator for the policy $\pi_\theta$:


\begin{equation}
    \nabla\pi_{\theta}\cdot\mathbbm{1}_{\text{stable}}\frac{1}{\pi_0}A,
\end{equation}

where $\mathbbm{1}_{\text{stable}}$ denotes a stability mask (e.g., clipping in GRPO) and $\pi_0$ represents the distribution used for importance sampling (e.g., $\pi_\theta$ for SFT and $\pi_{\theta_{\text{old}}}$ for GRPO). Despite its breadth, the original HPT framework omits Direct Preference Optimization (DPO), a pivotal tool in the post-training pipeline. Based on the connection between DPO and REINFORCE established in Sec.~\ref{sec:grpo}, we extend this framework to incorporate DPO, thereby completing the taxonomy.
Specifically, Tab.~\ref{tab:new_hpt} presents the extended framework for optimizing over a trajectory $\pi$ (a prompt-response pair), with our proposed additions highlighted in {\color{red}red}.


\begin{table}[ht]
    \centering
    \caption{The extended unified post-training framework. Here, $\pi_{\theta_{\text{old}}}$ refers to the policy prior to the current update; $\mathbbm{1}_{\text{clip}}$ denotes token-level clipping in GRPO/PPO, while $\mathbbm{1}_{\text{CIS-mask}}$ and $\mathbbm{1}_{\text{Seq-Clip}}$ refer to the specialized clipping strategies proposed in their respective papers. For FEST-GRPO, the gradient is defined as $\nabla\pi_\theta(\tau)\cdot\frac{A_1\mathbbm{1}_{\text{clip}}}{\pi_{\theta_{\text{old}}}(\tau)}+c\left(\frac{A_2\nabla\pi_\theta(\tau)}{\pi_\theta(\tau)}+\frac{A_2\nabla\pi_\theta(\tau)\mathbbm{1}_{\text{clip}}}{\pi_{\theta_{\text{old}}}}\right)$.}
    \begin{tabular}{cccc}
    \toprule
     Algorithm  & $\pi_0$ & Advantage $A$ & Gradient  \\
    \midrule
       SFT & $\pi_\theta$ & 1 & $\nabla\pi_\theta(\tau)\cdot\frac{1}{\pi_0(\tau)}$ \\
       PPO~\cite{schulman2017proximal} & $\pi_{\theta_{\text{old}}}$ & GAE~\cite{schulman2015high} & $\nabla\pi_\theta(\tau)\cdot\frac{A\mathbbm{1}_{\text{clip}}}{\pi_0(\tau)}$ \\
       GRPO~\cite{shao2024deepseekmath} & $\pi_{\theta_{\text{old}}}$ & $\frac{R(\tau_i)-\text{mean}(R(\tau_i))}{\text{std}(R(\tau_i))}$ & $\nabla\pi_\theta(\tau)\cdot\frac{A\mathbbm{1}_{\text{clip}}}{\pi_0(\tau)}$\\
       REINFORCE~\cite{ahmadian2024back} & $\pi_\theta$ & $\pm 1$ & $\nabla\pi_\theta(\tau)\cdot\frac{A}{\pi_0(\tau)}$ \\
       CISPO~\cite{chen2025minimax} & $\pi_{\theta_{\text{old}}}$ & $\frac{R(\tau_i)-\text{mean}(R(\tau_i))}{\text{std}(R(\tau_i))}$ & $\nabla\pi_\theta(\tau)\cdot\frac{A\mathbbm{1}_{\text{CIS-mask}}}{\pi_{0}(\tau)}$\\
       GSPO~\cite{zheng2025group} & $\pi_\theta\cdot(\frac{\pi_{\theta_{\text{old}}}}{\pi_\theta})^{1/|\tau|}$ & $\frac{R(\tau_i)-\text{mean}(R(\tau_i))}{\text{std}(R(\tau_i))}$ & $\nabla\pi_\theta(\tau)\frac{A\mathbbm{1}_{\text{Seq-Clip}}}{\pi_{0}(\tau)}$ \\
       SRFT~\cite{fu2025srft} & 1 & $\frac{R(\tau_i)-\text{mean}(R(\tau_i))}{\text{std}(R(\tau_i))}$ (incl. SFT trace) & $\nabla \pi_\theta(\tau)\cdot A$\\
       LUFFY~\cite{yan2025luffy} & 1 & $\frac{R(\tau_i)-\text{mean}(R(\tau_i))}{\text{std}(R(\tau_i))}$ (incl. SFT trace) & $\nabla\pi_{\theta}(\tau)\cdot A$\\
    \midrule
    \color{red} DPO~\cite{rafailov2023direct} & \color{red} $\pi_\theta$ & $\color{red}\pm \beta\cdot\sigma(\beta(r^--r^+))$ & \color{red}$\nabla\pi_\theta(\tau)\cdot \frac{A}{\pi_0(\tau)}$\\
    \color{red} FEST (ours) & \color{red} $\pi_\theta, \pi_{\theta_\text{old}}$ &\color{red} \makecell[t]{$A_1=R(\tau_i)-\text{mean}(R(\tau_i))$,\\ $A_2=\pm \beta\cdot\sigma(\beta(r^--r^+))$} & \color{red}\makecell[t]{ $\nabla\pi_\theta(\tau)\cdot\frac{A_1\mathbbm{1}_{\text{clip}}}{\pi_{\theta_{\text{old}}}(\tau)}+$\\$c\cdot\frac{ A_2\nabla\pi_\theta(\tau)}{\pi_\theta(\tau)}$}\\
    \bottomrule
    \end{tabular}
    
    \label{tab:new_hpt}
\end{table}

\section{Implementation Details}
\label{sec:implement}
\subsection{Pseudo-Code}
\label{sec:pseudo}

Alg.~\ref{alg:code} outlines the implementation of our proposed FEST algorithm. For simplicity, we omit the inner loop for multiple epochs per step, as all experiments in this study are conducted using a single epoch. We define the indicator function $\mathbb{I}[\text{condition}]$ as $1$ if the condition holds and $0$ otherwise.


\begin{algorithm}
\caption{FEST}\label{alg:code}
\begin{algorithmic}
\Require LLM with policy $\pi$, number of training steps $T$, batch size $B$, number of rollouts per question $N$, minibatch size $B_{\text{mini}}$, few-shot SFT dataset $D_E$, answer-only RL dataset $D_I$, hyperparameter $\beta_1, \beta_2,\beta_3$, loss coefficient $c>0$
\State Set $\pi_{\text{ref}}=\pi$
\For{$t=1,2,\dots,T$}
\State Sample questions $x^E_1,\dots,x^E_B$ from $D_E$ and $x^I_1,\dots,x^I_B$ from $D_I$
\State Set $\pi_{\theta_\text{old}}=\pi$
\For{$i=1,2,\dots,B$} 
\For{$j=1,2,\dots,N$}
\State Sample $y_{i,j}^E\sim \pi_{\theta_\text{old}}(\cdot|x_i^E)$, $y_{i,j}^I\sim \pi_{\theta_\text{old}}(\cdot|x_i^I)$
\EndFor
\State Use verifier to get reward $r^E_{i,j}(x_i^E,y^E_{i,j})$ and $r^I_{i,j}(x_i^I,y^I_{i,j})$
\State Calculate advantage on $I$: $A^I_{i,j}=r^I_{i,j}(x_i^I,y_{i,j}^I)-\frac{1}{N}\sum_{k=1}^Nr^I(x_i^I,y^I_{i,k})$
\State Calculate mask on $E$ for determining $\beta$: $M^{\text{correct}}_{i,j}=\mathbb{I}\left[r^E_{i,j}(x_i^E,y^E_{i,j})=1\right]$, $M^{\text{solvable}}_i=\mathbb{I}\left[\sum_{k=1}^Nr^E_{i,k}(x_i^E,y_{i,k}^E)>0\right]$
\EndFor

\For{$t_{\text{mini}}=1,2,\dots,2\times B\times N/B_{\text{mini}}$}
\State Sample $B_{\text{mini}}/2$ rollout from $D_E$ and $D_I$ respectively
\State Calculate $L_I$, the GRPO loss on $I$ using Eq.~\eqref{eq:grpo}
\State Determine $\beta$: $\beta^E_{i,j}=\left(\beta_1\cdot\left(1-M^{\text{solvable}}_i\right)+\beta_2\cdot M^{\text{solvable}}_i \right)\cdot\left[1-M^{\text{correct}}_{i,j}\right]+\beta_3\cdot\left[M^{\text{correct}}_{i,j}\right]$
\State Calculate $L_E$ using Eq.~\eqref{eq:fest-main} for FEST-DPO, or according to Sec.~\ref{sec:grpo} for FEST-GRPO
\State Update $\pi$ with loss $L=c\cdot L_E+L_I$
\EndFor
\EndFor
\end{algorithmic}
\end{algorithm}

\subsection{Hyperparameters}

Tab.~\ref{tab:hyperparam} lists the hyperparameters we used. Generally, hyperparameters are designed to match those of ReLIFT~\cite{ma2025relift}. The few-shot dataset is uniformly randomly sampled from the OpenR1-Math-46K-8192 dataset.

\begin{table}[ht]
\small
    \centering
    \caption{Experimental Hyperparameters. Due to occasional deadlock issues encountered with the standard \texttt{math-verify} library during distributed training, we utilized a modified reward verifier implementation from \url{https://github.com/RLHFlow/Reinforce-Ada}.}
    \begin{tabular}{lll}
    \toprule
       Hyperparameter  & Value & Note  \\
    \midrule
        Learning rate & 1e-5 decaying to 5e-6 & \\
        Learning rate scheduler & Cosine & \\
        Max question length & 1024 & in tokens; same as max response length \\
        Max response length & 8192 \\
        Batch size & 128 & questions sampled from each dataset \\ 
        Minibatch size & 512 & rollouts per gradient step \\
        Rollouts & 8 & samples per question \\
        KL regularizer & N/A & \\
        Entropy coefficient & 0.0001 & \\
        Grad clip & 1.0 & \\
        Max grad & 80 & step beyond this norm  will be discarded   \\
        GRPO clip ratio & (-0.2,0.3) & \\
        Training temperature & 1 & \\
        Eval temperature & 0.6 & \\
        $(\beta_1,\beta_2,\beta_3)$ & \makecell[t]{$(0.1, 0.01, 0.01)$ (FEST-DPO);\\ $(0.005,0.01,0.05)$ (FEST-GRPO)} & \\
        $c$ & 0.01; 1 (FEST-GRPO) & loss coefficient for $D_E$ \\
        Verifier & math-verify 0.8 & \\
        Max pos embedding & 16384 & \\
        RoPE~\cite{su2024roformer} $\theta$ & 40000 & following ReLIFT~\cite{ma2025relift}\\
        
    \bottomrule
    \end{tabular}
    
    \label{tab:hyperparam}
\end{table}

\subsection{Baselines}

We evaluate FEST against several state-of-the-art baselines. All implemented methods, including our own, are integrated into a unified codebase derived from ReLIFT\footnote{\url{https://github.com/TheRoadQaQ/ReLIFT}}. We utilize \texttt{verl} 0.4.0~\cite{sheng2024hybridflow} as our primary reinforcement learning backbone and \texttt{vllm} 0.8.4~\cite{kwon2023efficient} for rollout generation.\footnote{Due to the ARM-based architecture of our hardware, \texttt{vllm} was compiled from source.}


\begin{itemize}
    \item \textbf{HPT~\cite{lv2025hpt}:} HPT employs a selective training policy that applies SFT to samples where the average rollout reward falls below a specific threshold (set to $0$ for Qwen models) and RL otherwise. Our results are based on an implementation with a learning rate of $5 \times 10^{-6}$ (following their work) and generation configurations aligned with the original study. Notably, in our few-shot setting, we found that restricting HPT to SFT-only on the 128-shot gold dataset $D_E$ yielded better performance than combining it with RL (see Tab.~\ref{tab:main}).
    

    \item \textbf{ReLIFT~\cite{ma2025relift}:} Similar to HPT, ReLIFT identifies samples with $0$ reward but stores them in a buffer, executing a standalone SFT step once the buffer reaches a threshold size. We implement ReLIFT with a learning rate of $5 \times 10^{-6}$, an adjustment from the $1 \times 10^{-6}$ used in the original paper and the Unify-Post-Training repository, as we observed that this higher rate significantly enhanced performance in our experiments.
    

    \item \textbf{LUFFY~\cite{yan2025luffy}:} LUFFY interleaves SFT and RL gradients during every update step by treating SFT demonstrations as rollouts within a regularized importance sampling framework. Consistent with our other evaluations, we found that increasing the learning rate to $5 \times 10^{-6}$ improved results, though further increases to $1 \times 10^{-5}$ induced training instability. We maintain consistency in online sample counts by using 8 online rollouts and 1 offline SFT sample per update.
    
    
    \item \textbf{CHORD~\cite{zhang2025onpolicy}:} CHORD conducts SFT on the SFT dataset and RL on the answer-only dataset simultaneously with decaying weight. There are two variants: CHORD-$\mu$ naively applies a decaying weight on SFT, while CHORD-$\phi$ applies a weight of $\phi(x)=p_x(1-p_x)$ on each token, where $p_x$ is the probability of the sampled token $x$. Note, while CHORD provides code base, it is part of a highly integrated framework called Trinity-RFT~\cite{trinity-rft}, and limits training data to problems with integer answer as it lacks a comprehensive verifier like \texttt{math-verify}~\cite{math_verify2025}. Thus, the code base is not directly usable and we implement it in our code base. As CHORD-$\phi$ outperforms CHORD-$\mu$ in their paper, we implement and test CHORD-$\phi$ with a learning rate of 5e-6. 

    \item \textbf{SRFT~\cite{fu2025srft}:} Similar to CHORD, SRFT mixes RL and SFT loss from both datasets for every gradient step with entropy-based weights. As SRFT does not release code (the link to their repository \url{https://anonymous.4open.science/w/SRFT2025/} has expired) but has available checkpoints based on Qwen2.5-Math-1.5B, we directly download their checkpoint (\url{https://huggingface.co/Yuqian-Fu/SRFT-Qwen2.5-Math-1.5B}) and test the results, which is far worse than our method. The code base of HPT also implemented SRFT; however, SRFT training constantly collapse with their implementation. Based on results from ReLIFT and LUFFY, we hypothesize that the performance of the official checkpoint could be limited due to conservative learning rate.

    \item \textbf{MIFO~\cite{yuan2025mitigating}:} MIFO utilizes a buffer-based SFT approach similar to ReLIFT but selectively updates model parameters that exhibit high sensitivity during RL. In the absence of an open-source codebase or public checkpoints, we cite the performance metrics directly from the original paper for the Qwen2.5-Math-1.5B base model.
    
\end{itemize}

\section{More Experimental Results}
\label{sec:more_results}

\subsection{Supplementary Results for Methodology}
\label{sec:complement}

\begin{figure}
    \centering
    \begin{minipage}[c]{0.45\linewidth}
    \subfigure[SFT trajectory ratio]{\includegraphics[width=\linewidth]{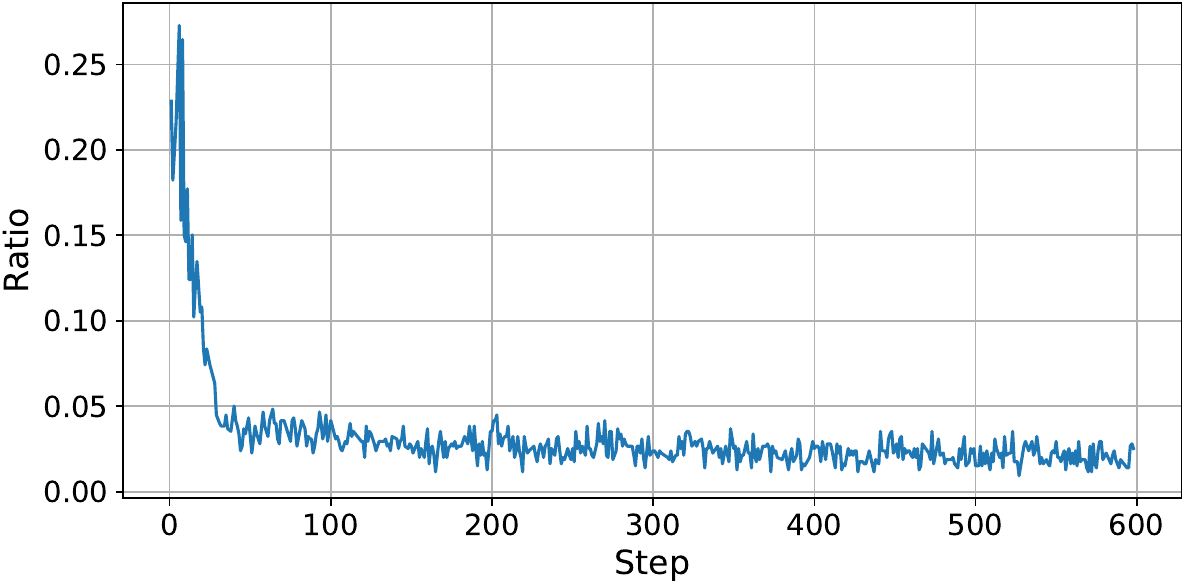}}
    \end{minipage}
    \begin{minipage}[c]{0.45\linewidth}
    \subfigure[Test set performance]{\includegraphics[width=\linewidth]{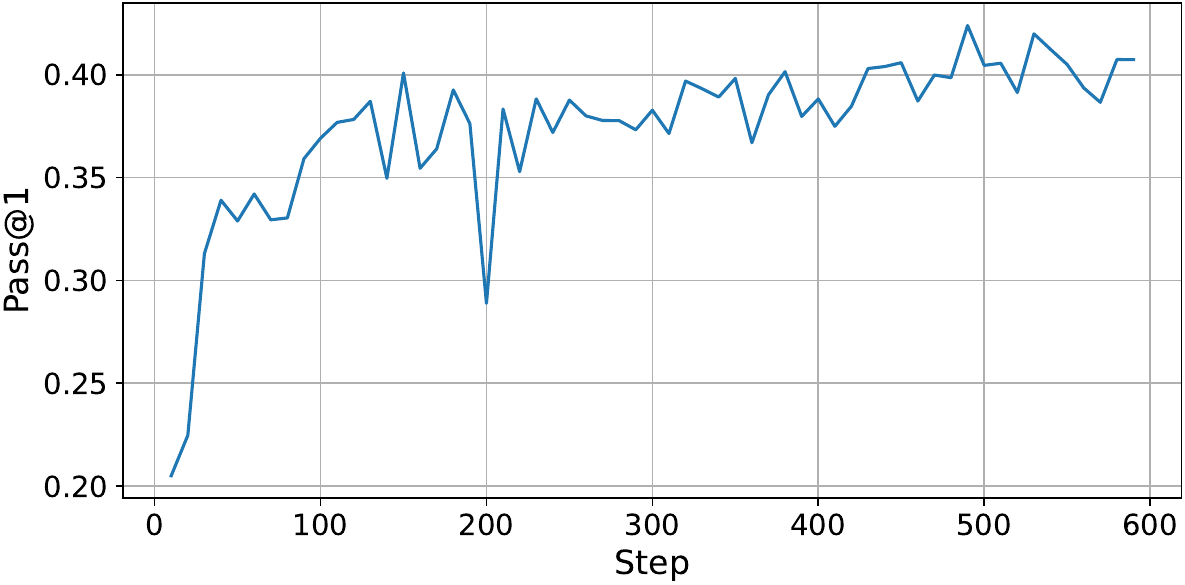}}
    \end{minipage}
    \caption{Evolution of SFT trajectory ratios and corresponding Pass@1 performance during HPT replication on the full dataset. The ratio of expert trajectories declines to approximately 2\% of total rollouts, yet performance metrics consistently trend upward. This observation provides empirical justification for the use of a decaying weight strategy in the SFT objective during later training phases.}
    \label{fig:hpt_ratio}
\end{figure}

\textbf{HPT Data Ratio Dynamics.} We replicated the HPT framework using the official implementation~\cite{lv2025hpt} and observed that the proportion of SFT trajectories decreases significantly as training progresses, reaching as low as 2\% in the terminal stages (see Fig.~\ref{fig:hpt_ratio}). Notably, test set performance continues to improve even as the SFT signal diminishes. This empirical finding reinforces our hypothesis that the SFT component in demonstration-guided RLVR—regardless of whether the setting is few-shot or large-scale—should follow a decaying schedule. This principle is further corroborated by the design of CHORD~\cite{zhang2025onpolicy}, which incorporates similar decaying mechanisms in both its variants.


\begin{figure}[ht]
    \centering
    \includegraphics[width=0.5\linewidth]{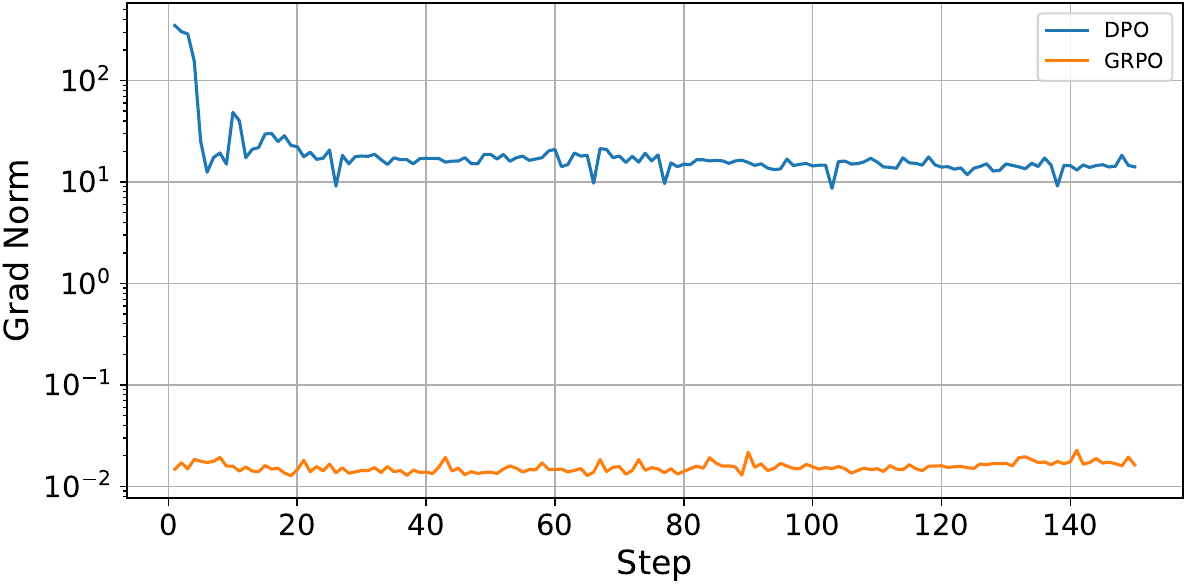}
    \caption{Gradient norm comparison between DPO and GRPO objectives when applied independently. The sequence-based DPO objective yields gradients multiple orders of magnitude larger than the token-based GRPO. FEST-GRPO harmonizes these scales to ensure stable joint optimization.}
    \label{fig:grad}
\end{figure}

\textbf{Gradient Mismatch Analysis.} As discussed in Sec.~\ref{sec:grpo}, a fundamental structural discrepancy exists between the sequence-level objective of DPO and the token-level objective of GRPO. In DPO, the log-probabilities of the entire sequence are aggregated within the log-sigmoid function, whereas GRPO computes the average gradient across individual tokens. This difference, exacerbated by the extended reasoning traces characteristic of RLVR—which are significantly longer than typical RLHF sequences—results in a severe magnitude mismatch between the respective gradients. Specifically, as illustrated in Fig.~\ref{fig:grad}, the norm of the DPO gradient typically ranges between $10^1$ and $10^2$, while the GRPO gradient norm consistently remains below $0.1$. Our proposed FEST-GRPO variant naturally resolves this imbalance by unifying both objectives within a consistent token-level framework, thereby eliminating the need for exhaustive hyperparameter search for the coefficient $c$.


\subsection{HPT-G and ReLIFT-G}
\label{sec:withgold}

\begin{figure}[ht]
    \centering
    \begin{minipage}[c]{0.45\linewidth}
    \subfigure[Reward (Avg@8) performance on $D_I$]{\includegraphics[width=\linewidth]{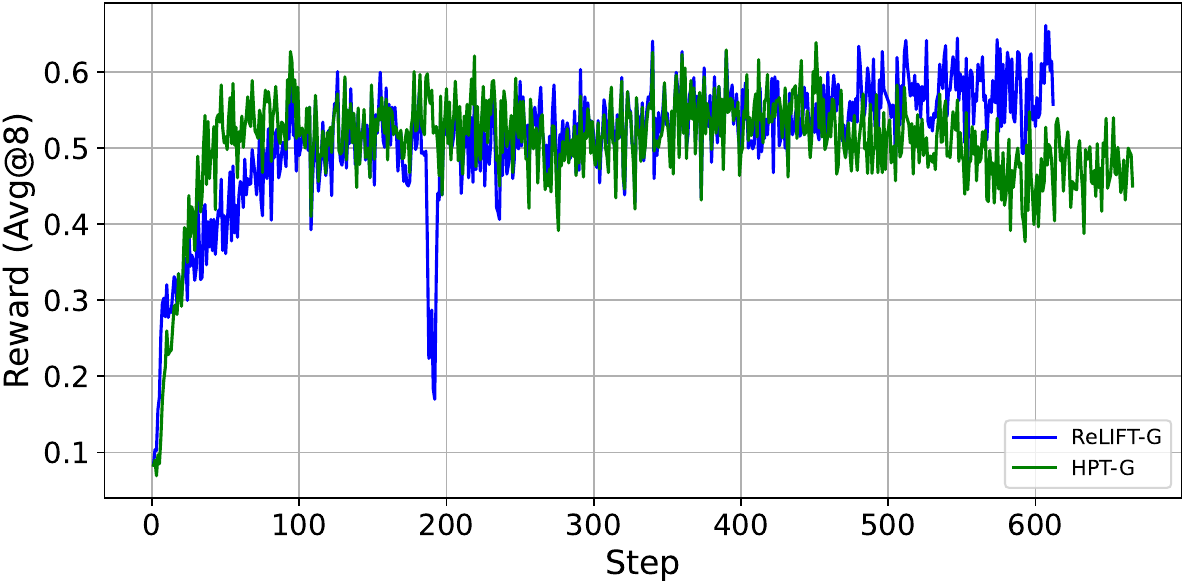}}
    \end{minipage}
    \begin{minipage}[c]{0.45\linewidth}
    \subfigure[Reward (Avg@8) performance on $D_E$]{\includegraphics[width=\linewidth]{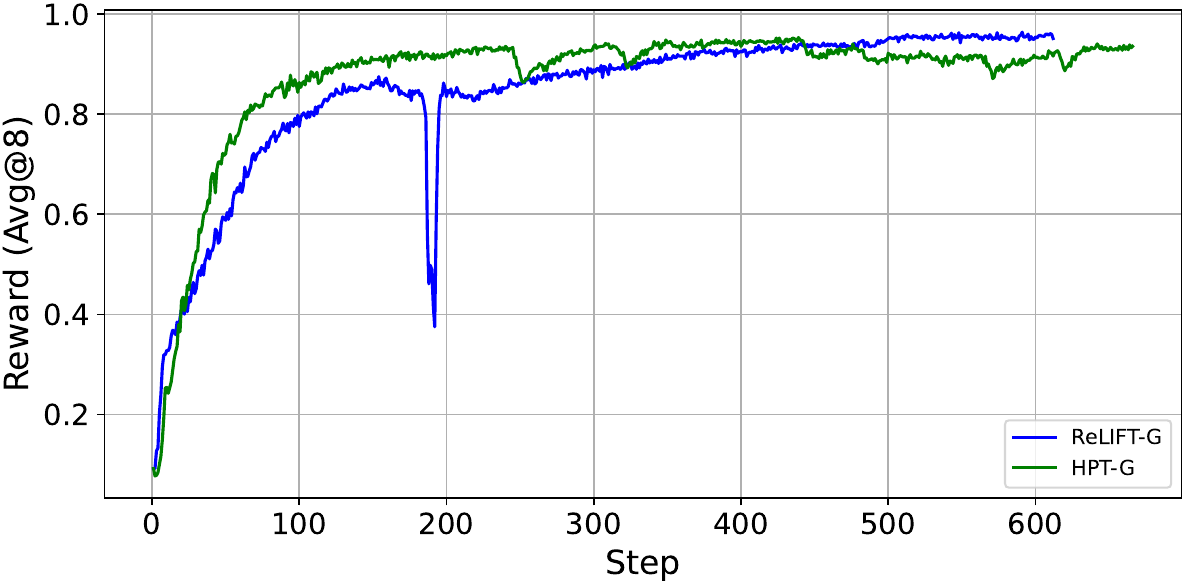}}
    \end{minipage}
    \caption{Training set accuracy profiles for ReLIFT-G and HPT-G. Both variants exhibit significant training instability, characterized by sudden and simultaneous performance drops.}
    \label{fig:withgold}
\end{figure}

Fig.~\ref{fig:withgold} characterizes the training dynamics of HPT-G and ReLIFT-G, specifically focusing on the impact of applying reinforcement learning to the few-shot gold dataset $D_E$. As evidenced by the plots, both methods experience acute performance degradation in the middle of the training process across both $D_E$ and the RL dataset $D_I$. We hypothesize that this instability is driven by abrupt distribution shifts: the SFT gradients introduce localized updates that conflict with a policy already heavily optimized (and potentially overfitted) by RL on the limited samples of $D_E$. This behavioral collapse underscores the difficulty of naively mixing objectives on sparse data and highlights the importance of the decaying weight strategy employed by FEST to maintain optimization stability.

\subsection{Analysis on \texorpdfstring{$\beta$}{beta}}
\label{sec:ablation_hyperparam}

To investigate how the choice of $\beta$ influences performance, we conduct a sensitivity analysis on the hyperparameter $\beta$ within the FEST-DPO framework. Our setting diverges from standard DPO applications in two fundamental ways: \textbf{extended Chain-of-Thought (CoT) reasoning} and the \textbf{few-shot regime}. The former involves reasoning traces of up to 8,192 tokens—over a magnitude longer~\cite{bai2022training} than typical RLHF sequences—leading to significantly larger cumulative log-probabilities. The latter necessitates repeated training epochs over a minuscule dataset, which further drives the discrepancy between $\log\pi_\theta(y^+|x)$ and $\log\pi_\theta(y^-|x)$. Consequently, the log-ratio difference reaches magnitudes far exceeding those encountered in traditional alignment tasks.

\begin{figure}[ht]
    \centering
   \begin{minipage}[c]{0.47\linewidth}
    \subfigure[Average $z$]{\includegraphics[width=\linewidth]{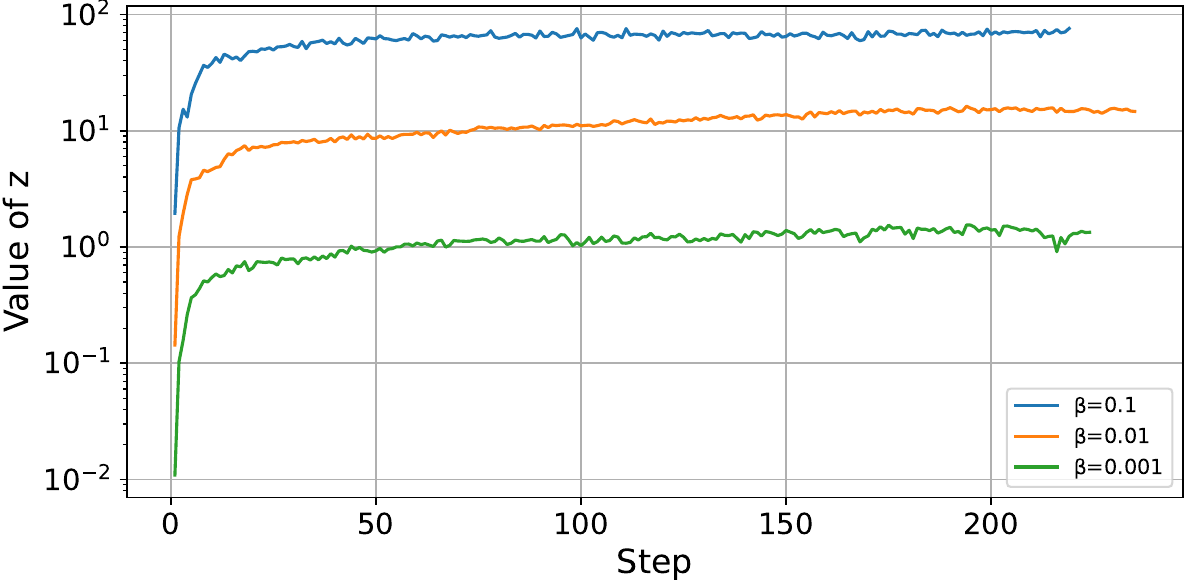}}
    \end{minipage}
    \begin{minipage}[c]{0.47\linewidth}
    \subfigure[Minimum $z$]{\includegraphics[width=\linewidth]{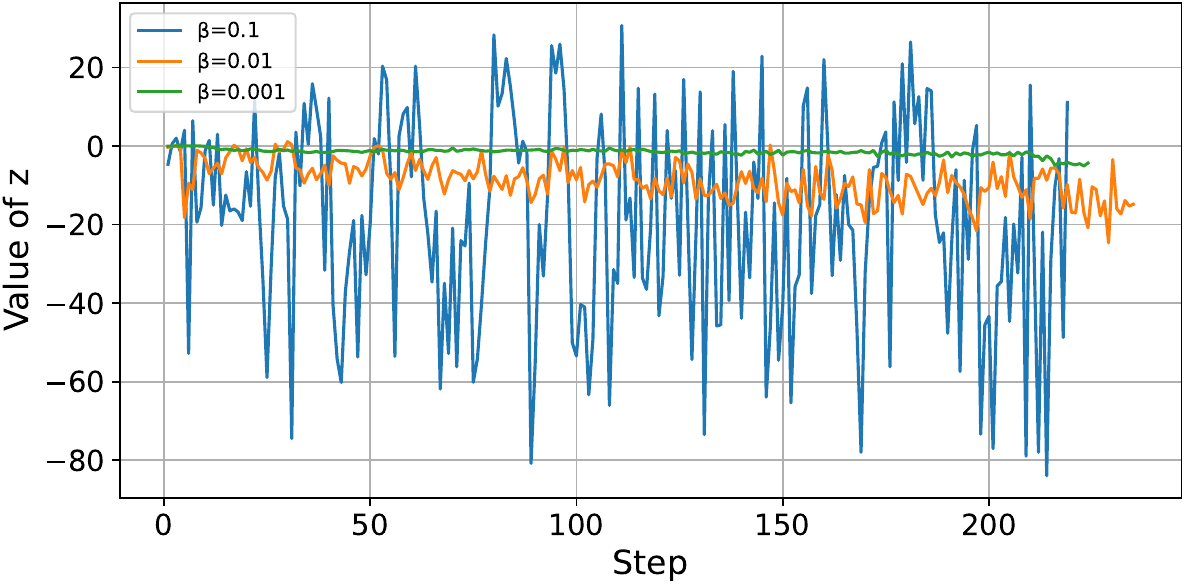}}
    \end{minipage}
    \caption{Implicit advantage $z$ as a function of $\beta$. Panel (a) shows that average $z$ scales approximately linearly with $\beta$, i.e., the log-ratio difference is nearly consistent and our assumption holds for $\beta\in[0.001,0.1]$. Panel (b) reveals that higher $\beta$ values result in a wider distribution of $z$, where a few examples with very low $z$ receive intense ``switch-like'' signals. Conversely, lower $\beta$ values yield a more concentrated $z$ distribution, resulting in a more constant and stable learning signal.}
    
    \label{fig:ablation_z}
\end{figure}

To understand how these factors dictate the optimal $\beta$, we re-examine the DPO gradient in Eq.~\eqref{eq:grad}, where the weight applied to the score gradient is $\beta \cdot \sigma(r^- - r^+)$. Adopting DPO’s terminology for ``implicit reward,'' we define the \textit{implicit advantage} $z$ for a given pair as:

\begin{equation}
    z=\beta\cdot\left(\log\frac{\pi_\theta(y^+|x)}{\pi_{\text{ref}}(y^+|x)}-\log\frac{\pi_\theta(y^-|x)}{\pi_{\text{ref}}(y^-|x)}\right)=\beta\cdot\Delta,
\end{equation}

and the resulting gradient coefficient is $w = \beta \cdot \sigma(-z) = \frac{\beta}{1 + e^z}$. Because $z$ is typically very large in our reasoning tasks (see Fig.~\ref{fig:ablation_z} (a)), this coefficient is dominated by the exponential term rather than the linear factor $\beta$. Paradoxically, decreasing $\beta$ can lead to a much stronger learning signal: assuming a fixed log-ratio difference $\Delta \gg 1$, the gradient coefficient is approximately $\beta e^{-z}$. In this case, reducing $\beta$ to $0.1\beta$ causes the coefficient to scale by a factor of approximately $0.1 e^{0.9z}$. The balance threshold for this scaling is $z \approx 2.56$, which, according to Fig.~\ref{fig:ablation_z} (a), corresponds to a $\beta$ value slightly above $0.001$. Thus, counterintuitively, \textbf{when $\beta>0.001$, smaller $\beta$ values provide stronger and more persistent learning signals}, necessitating the use of values significantly lower than the standard $0.1$--$0.2$ range~\cite{rafailov2023direct} (e.g., $0.001$--$0.1$).


However, according to Fig.~\ref{fig:ablation_z} (b), on the other hand, the absolute value of the minimum value of $z$ also greatly increases. Such phenomenon indicates that the learning signal strength is not monotonic with respect to $\beta$, but instead forms a ``switch-constant'' dichotomy:

\begin{itemize}
    \item With larger $\beta$, the learning is closer to a ``switch'', where the signal strength will be exceptionally strong at the beginning of training or in some corner cases where $y^+$ is less likely to be sampled than $y^-$, but quickly decrease to almost $0$;

    \item With smaller $\beta$, the learning signal is strong (instead of weak in usual belief~\cite{rafailov2023direct}) and constant.
\end{itemize}

Based on these findings, we conducted an ablation study on $\beta$ for FEST-DPO, with results summarized in Tab.~\ref{tab:ablation-z}. The results show that while our method is robust across a range of hyperparameters, values that are too far from the balance point (either too large or too small) can degrade performance. We selected the optimal configuration for FEST-DPO. For FEST-GRPO, we heuristically chose a set of hyperparameters such that $z$ predominantly falls between $1$ and $10$, maintaining the signal near the balance point of $z \approx 2.56$.


\begin{table}[ht]
    \centering
    \caption{Test set performance (Avg@8) with FEST-DPO using different $\beta$. Generally, we find that our algorithm works well with $\beta$ between 0.001 and 0.1, but suffers from performance loss if all hyperparameters are too large (close to 0.1) or too small (close to 0.001).}
    \begin{tabular}{cc}
    \toprule
      $(\beta_1,\beta_2,\beta_3)$  & Avg@8  \\
    \midrule
        (0.1, 0.1, 0.1) & 40.22$\pm$1.19 \\
         (0.1, 0.1, 0.05) & 41.51$\pm$1.32 \\
         (0.1, 0.1, 0) & 41.76$\pm$1.02 \\
        (0.1, 0.01, 0.01) & 41.99$\pm$1.24 \\
        (0.01, 0.01, 0.001) & 41.33$\pm$2.16 \\
        (0.001, 0.001, 0.001) & 40.24$\pm$1.22 \\
    \bottomrule
    \end{tabular}
    
    \label{tab:ablation-z}
\end{table}

\subsection{Training Curves}
\label{sec:curves}

\begin{figure}
    \centering
    \begin{minipage}[c]{0.47\linewidth}
    \subfigure[$D_E$ (few-shot SFT training set) accuracy]{\includegraphics[width=\linewidth]{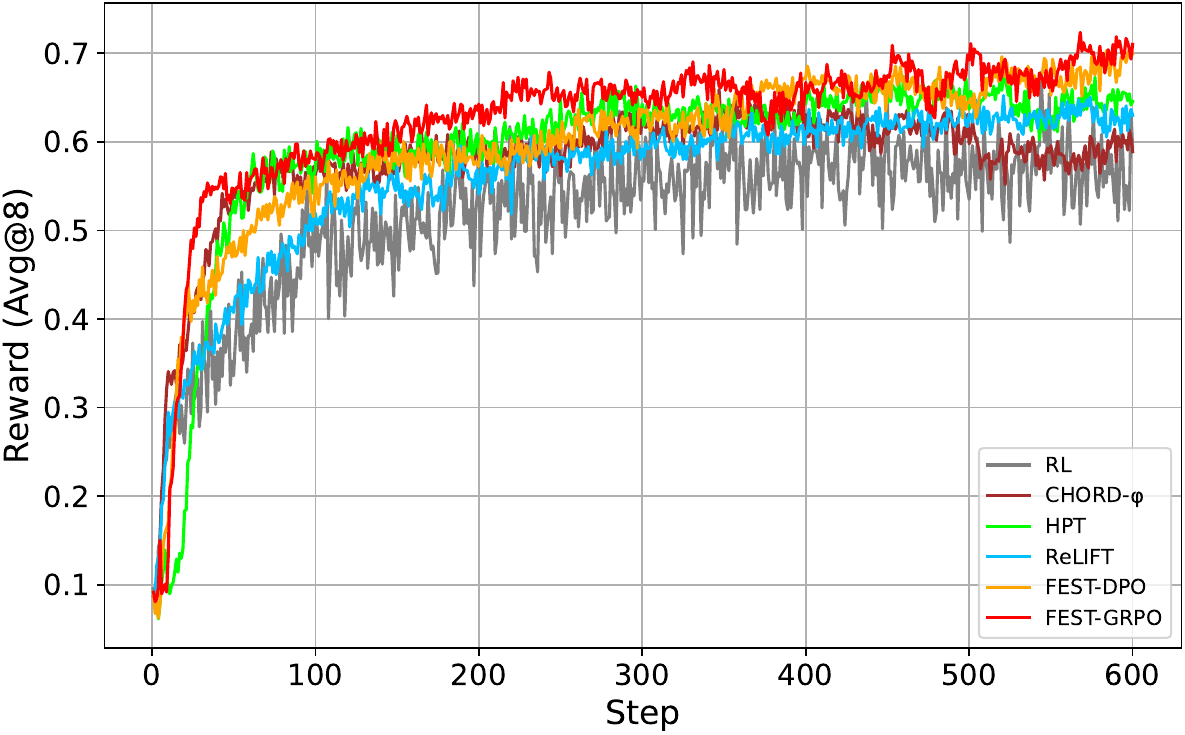}}
    \end{minipage}
     \begin{minipage}[c]{0.47\linewidth}
    \subfigure[zoomed-in curve (100--600 step)]{\includegraphics[width=\linewidth]{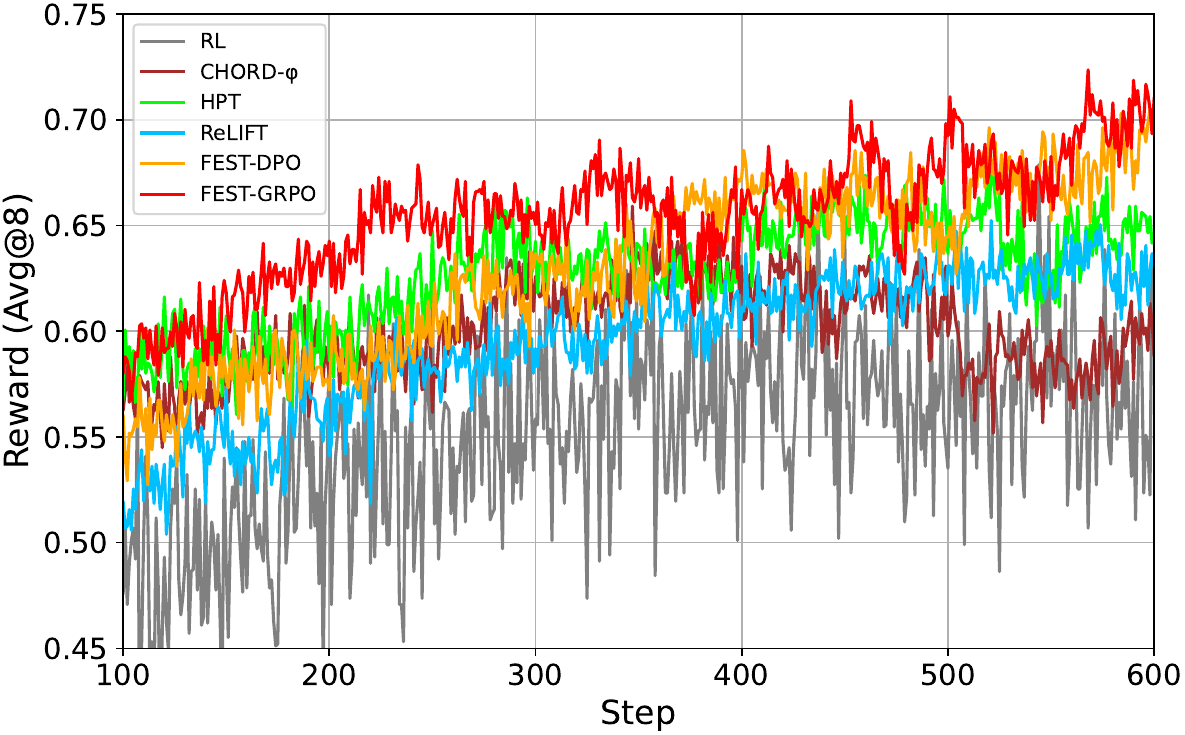}}
    \end{minipage}
    \caption{Reward curves on $D_E$ during training. Results for the ``higher group'' (LUFFY, ReLIFT-G, HPT-G, and RL-G) are excluded from these plots as their direct RL training on $D_E$ leads to near 100\% training accuracy, masking meaningful comparative dynamics. The performance implications of this overfitting are discussed in Appendix~\ref{sec:complement} and Tab.~\ref{tab:pass8}.}
    \label{fig:curves-gold}
\end{figure}

\begin{figure}
    \centering
    \begin{minipage}[c]{0.47\linewidth}
    \subfigure[$D_I$ (RL training set) accuracy]{\includegraphics[width=\linewidth]{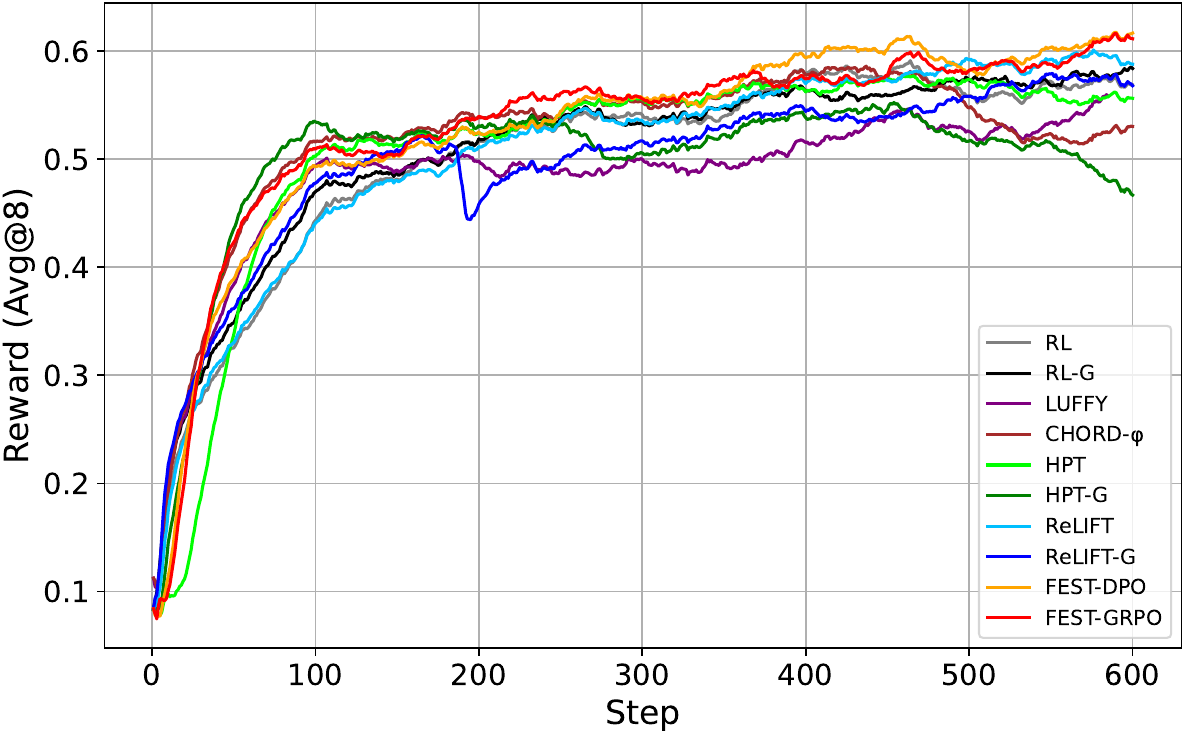}}
    \end{minipage}
    \begin{minipage}[c]{0.47\linewidth}
    \subfigure[zoomed-in curve (100--600 step)]{\includegraphics[width=\linewidth]{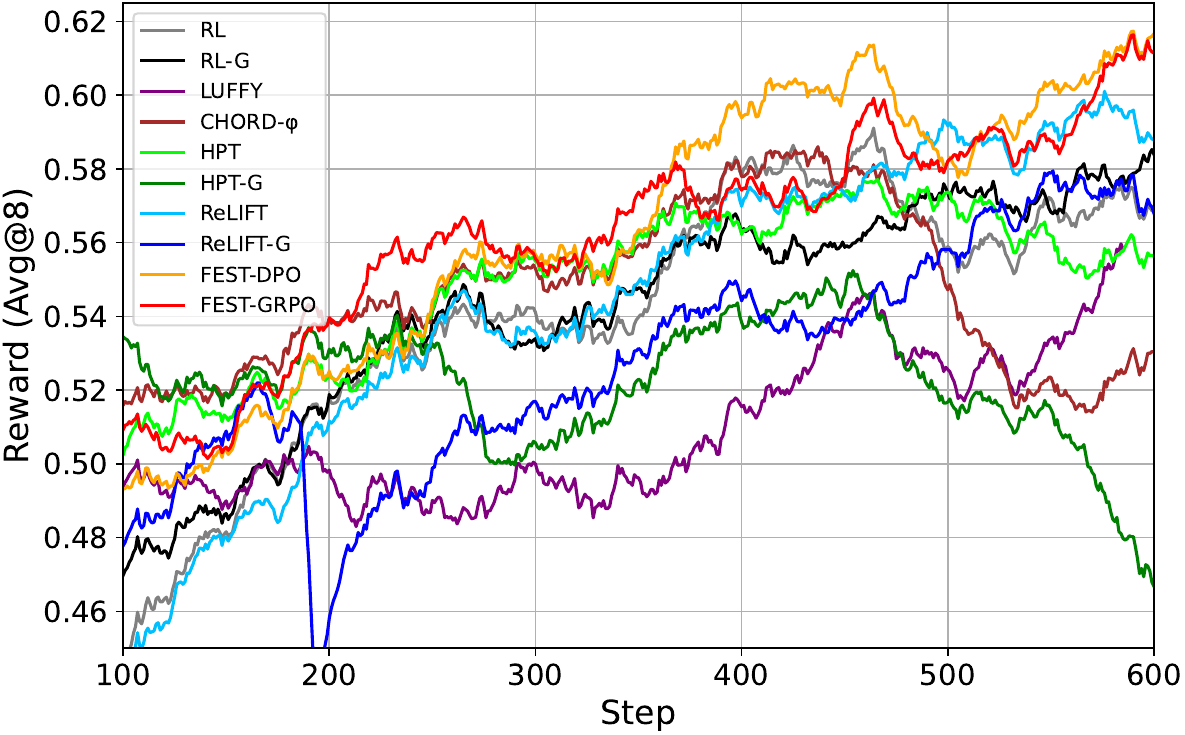}}
    \end{minipage}
    \caption{Smoothed reward curves on $D_I$ utilizing a time-weighted exponential moving average.}
    \label{fig:curves-rl}
\end{figure}

\begin{figure}
    \centering
    \begin{minipage}[c]{0.47\linewidth}
    \subfigure[$D_I$ (RL training set) accuracy]{\includegraphics[width=\linewidth]{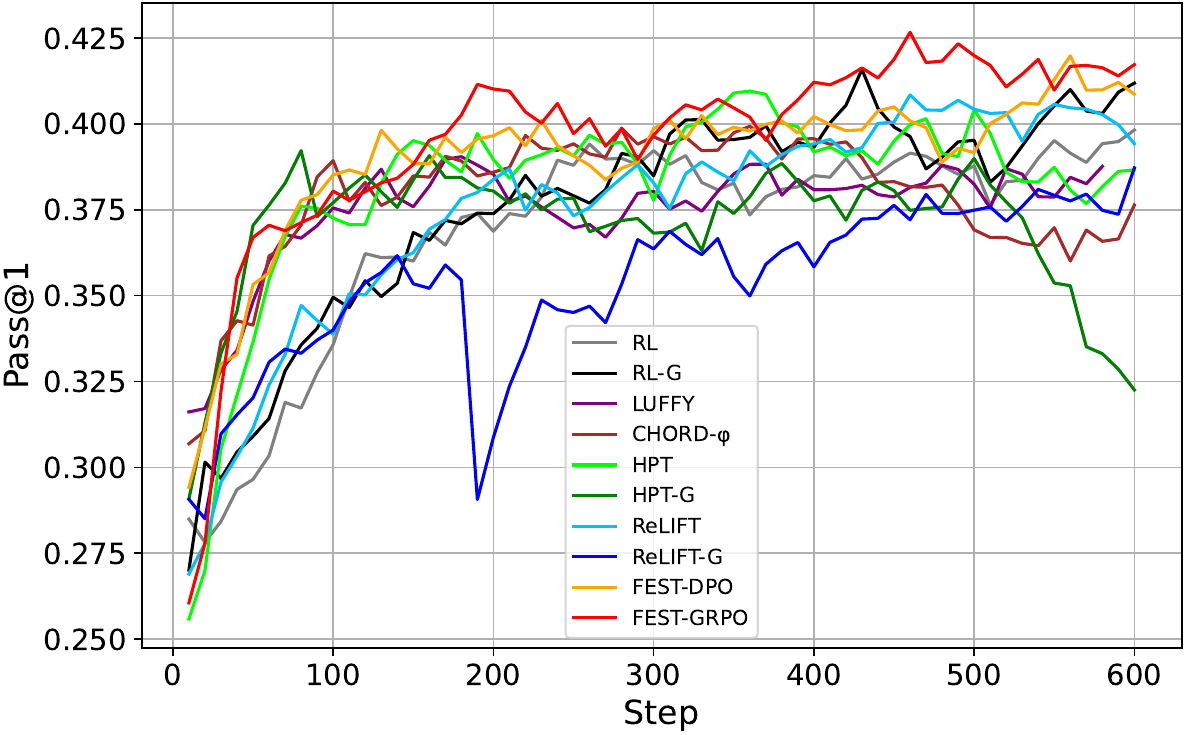}}
    \end{minipage}
     \begin{minipage}[c]{0.47\linewidth}
    \subfigure[zoomed-in curve (100--600 step)]{\includegraphics[width=\linewidth]{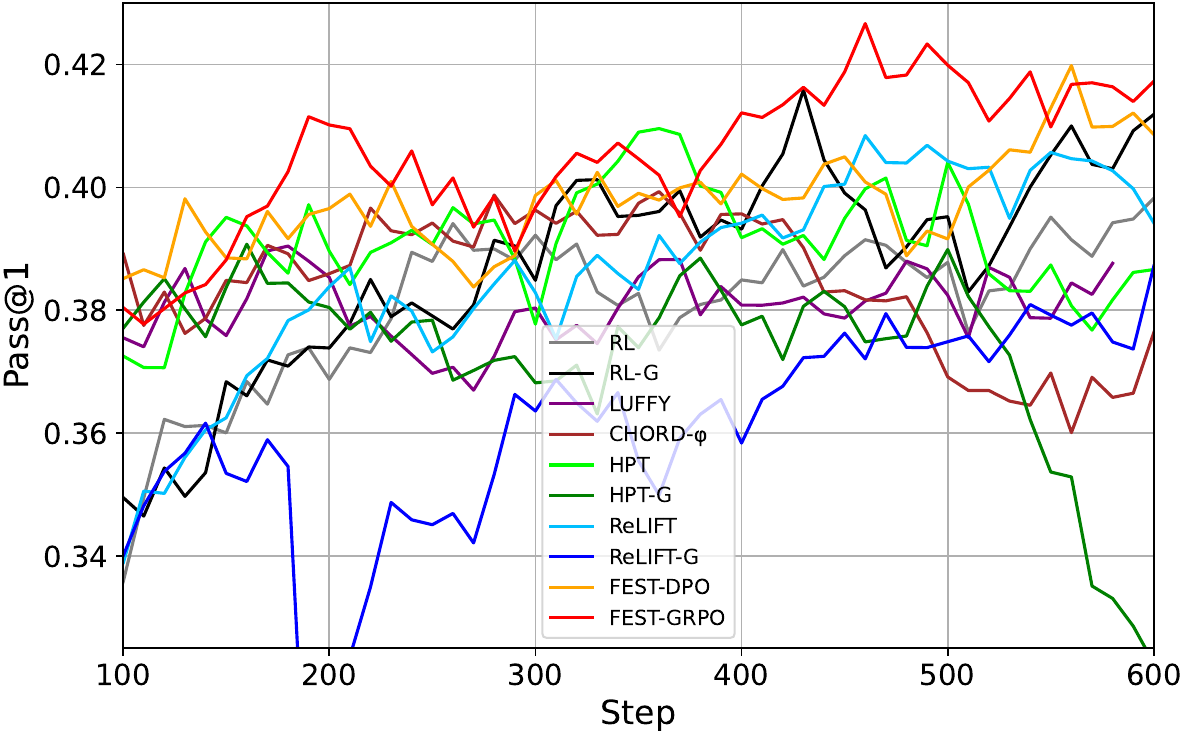}}
    \end{minipage}
    \caption{Smoothed Pass@1 performance on the test set through the training process.}
    \label{fig:curves-test}
\end{figure}

\textbf{Primary Training Results (Sec.~\ref{sec:main}).} Figures~\ref{fig:curves-gold}, \ref{fig:curves-rl}, and \ref{fig:curves-test} illustrate the accuracy profiles on the few-shot SFT dataset $D_E$, the RL dataset $D_I$, and the test set, respectively. For enhanced readability, we provide zoomed-in views focusing on the 100--600 step interval. The test set performance is smoothed due to its high variance nature (to prevent interrupting the training process, we use Pass@1 for evaluation during training). We identify three critical trends in these dynamics:


\begin{itemize}
    \item \textbf{Stratification of Reward Profiles on $D_E$:} The accuracy curves on $D_E$ bifurcate into two distinct groups. The higher-performing group (ReLIFT-G, HPT-G, RL-G, and LUFFY) consists of methods that directly apply RL to the gold dataset. Within this group, ReLIFT-G and HPT-G exhibit significant instability compared to RL-G and LUFFY, likely due to the high SFT weighting. Conversely, LUFFY maintains better stability through regularized importance sampling; however, this stability is sensitive to the learning rate, as we observed instability when LUFFY was scaled to our specific configuration.

    \item \textbf{Convergence vs. Overfitting:} While LUFFY, CHORD-$\phi$, and HPT exhibit rapid initial convergence comparable to FEST, they eventually suffer from performance stagnation on the test set, indicating a susceptibility to overfitting. In contrast, under few-shot constraints, ReLIFT closely mirrors vanilla RL dynamics because its SFT triggering mechanism—averaging only once every 20 RL steps—is too infrequent to provide meaningful expert guidance.

    \item \textbf{FEST Superiority:} Across the $D_E$ (lower group), $D_I$, and test set metrics, both FEST-DPO and FEST-GRPO consistently maintain a substantial performance margin over all baseline methods throughout the training process.
\end{itemize}


\begin{figure}
    \centering
    \begin{minipage}[c]{0.32\linewidth}
    \subfigure[$D_I$ (RL training set) accuracy]{\includegraphics[width=\linewidth]{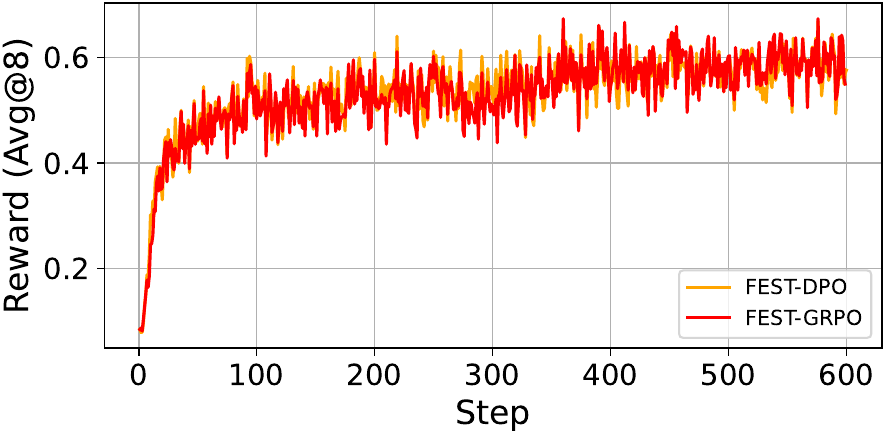}}
    \end{minipage}
    \begin{minipage}[c]{0.32\linewidth}
    \subfigure[$D_E$ (Few-shot SFT set) accuracy]{\includegraphics[width=\linewidth]{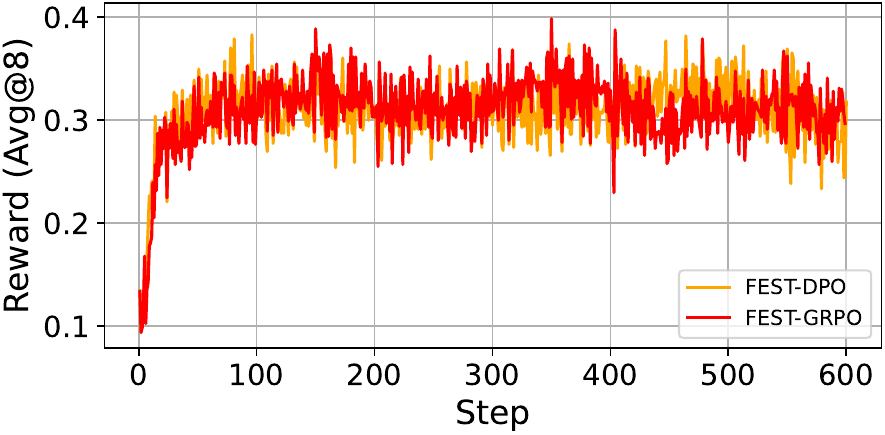}}
    \end{minipage}
    \begin{minipage}[c]{0.32\linewidth}
    \subfigure[Pass@1 on test set]{\includegraphics[width=\linewidth]{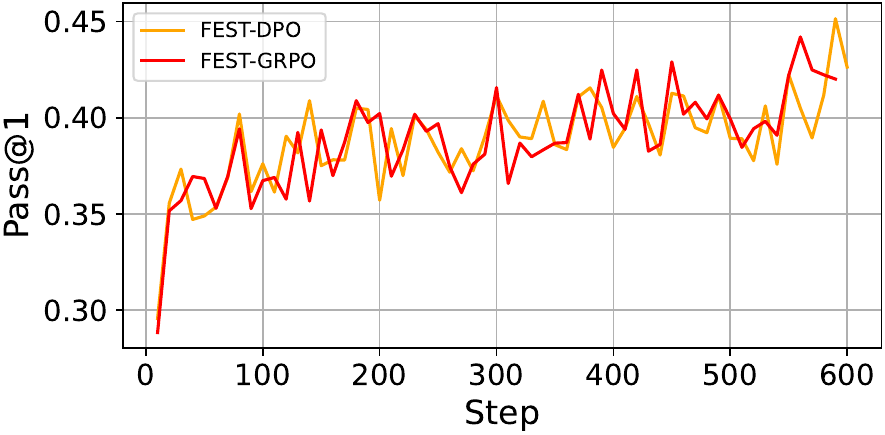}}
    \end{minipage}
    \caption{Training and test set performance curves for the LIMOv2-8192 experiment described in Sec.~\ref{sec:choice}.}
    \label{fig:curves_limo}
\end{figure}

\textbf{Dynamics on LIMOv2-8192.} Fig.~\ref{fig:curves_limo} presents the training dynamics for the LIMOv2-8192 experiment (Sec.~\ref{sec:choice}). While performance on $D_I$ and the test set shows continuous growth, accuracy on $D_E$ remains relatively stationary. We attribute this to the substantially higher reasoning difficulty of the LIMOv2 dataset compared to standard splits. Nevertheless, the framework successfully extracts performance gains from this specialized few-shot dataset, demonstrating robust cross-dataset applicability.

\subsection{SFT and SPIN}
\label{sec:sftspin}

\begin{figure}
    \centering
    \includegraphics[width=0.5\linewidth]{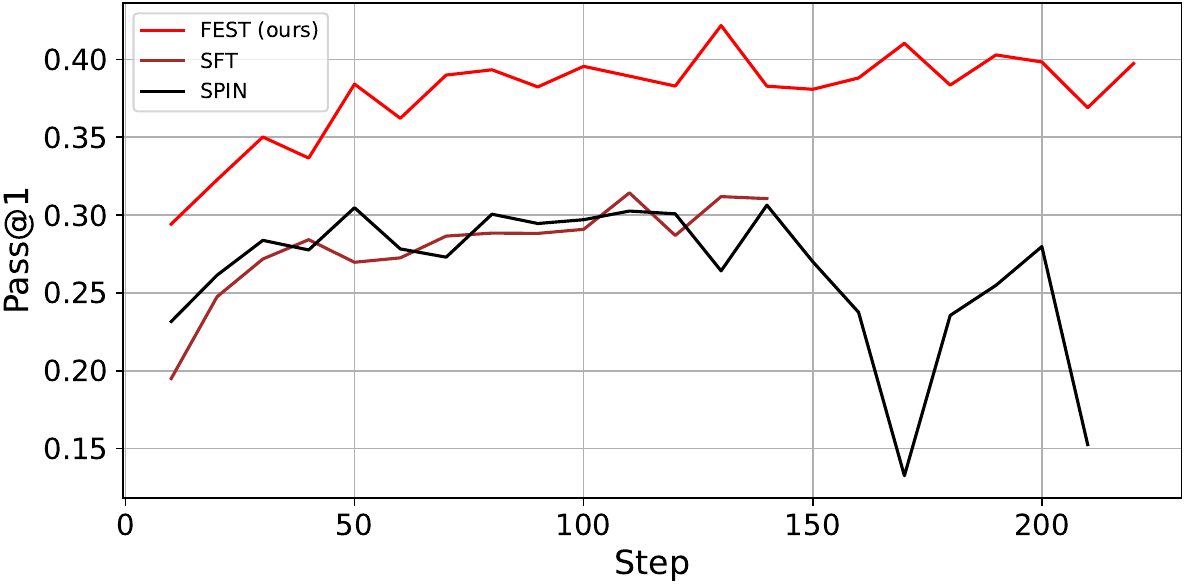}
    \caption{Test set performance comparison between FEST, standalone SFT, and SPIN. Our method significantly outperforms both baselines. SPIN, in particular, demonstrates a characteristic overfitting trajectory on the few-shot dataset, with performance peaking early and then rapidly declining. Due to this divergence and limited computational resources, these baselines were not extended to the full training duration or reported in the main paper.}
    \label{fig:sftandspin}
\end{figure}

In Sec.~\ref{sec:main}, we omit the formal performance report for standalone SFT and SPIN~\cite{chen2024self} on the 128-shot dataset. This exclusion is due to the inherent inability of these methods to generalize within extreme few-shot regimes without the benefit of an auxiliary RL objective on $D_I$. Specifically, both methods exhibit rapid and severe overfitting from the earliest stages of optimization. Consequently, these training runs were terminated prior to the 600-step mark to prioritize computational resource allocation. Figure~\ref{fig:sftandspin} illustrates these dynamics.


\section{Computational Resources}
\label{sec:resources}

All experiments were conducted on a single compute node equipped with two NVIDIA GH200 (96GB) Grace Hopper Superchips, 64 ARM-based CPU cores, and 256GB of system memory. A standard training run of 600 steps requires approximately 4--5 days of wall-clock time to complete, while an evaluation across our benchmark suite (utilizing 8 rollouts per question) takes approximately 30 minutes.


\section{Licenses}
\label{sec:license}

The licensing information for the datasets, benchmarks, and baseline models utilized in this study is summarized in Tab.~\ref{tab:license}. While several assets lack explicit licensing documentation, they are recognized as standard resources within the research community and have been extensively utilized in prior work~\cite{lv2025hpt,ma2025relift,yan2025luffy}.


\begin{table}[ht]
    \centering
    \caption{Licenses of the assets utilized in this work. Entries marked as ``N/A'' indicate that no explicit license was provided by the original authors or repositories.}
    \begin{tabular}{lcc}
    \toprule
    Asset Name & Category & License \\
    \midrule
    Qwen2.5-Math-1.5B & base model & Apache-2.0 \\
 OpenR1-Math-46K-8192 & training set & MIT  \\
       LIMOv2  & training set & Apache-2.0 \\
       AMC23 & benchmark & N/A\\
       AIME24 & benchmark & Apache-2.0 \\
       AIME25 & benchmark & Apache-2.0\\
       OlympiadBench & benchmark & N/A \\
       MATH-500 & benchmark & N/A\\
       Minerva & benchmark & N/A \\
       MMLU-Pro & benchmark & Apache-2.0 \\
       ReLIFT & code base & N/A\\
       HPT & code base & MIT \\
       CHORD & code base & Apache-2.0 \\
       LUFFY & code base & N/A \\
       SRFT & checkpoint & MIT \\
    \bottomrule
    \end{tabular}
    
    \label{tab:license}
\end{table}





\end{document}